\documentclass[10pt,twocolumn,letterpaper]{article}

\usepackage[pagenumbers]{cvpr} 

\usepackage{amsmath,amssymb,amsthm,bm,booktabs,multirow,xspace}
\usepackage{graphicx}

\DeclareMathOperator{\diag}{diag}
\DeclareMathOperator{\Cov}{Cov}
\DeclareMathOperator{\Var}{Var}

\newcommand{\E}{\mathbb{E}}
\newcommand{\R}{\mathbb{R}}
\newcommand{\x}[1]{\bm{#1}}
\newcommand{\xt}{\x{x}_t}
\newcommand{\xo}{\x{x}_0}
\newcommand{\xhat}{\hat{\x{x}}_0}
\newcommand{\one}{\mathbf{1}}
\newtheorem{definition}{Definition}
\newtheorem{proposition}{Proposition}
\newtheorem{example}{Example}

\definecolor{cvprblue}{rgb}{0.21,0.49,0.74}
\usepackage[pagebackref,breaklinks,colorlinks,allcolors=cvprblue]{hyperref}

\def\confName{CVPR}
\def\confYear{2027}

\title{JaRS: Contrast-Free Localization of Memorized Regions\\in Diffusion Models via Jacobian Row Sums}

\author{
    Saisuchith Mahajan, Aditey Nandan, Rohit Singh, Nishanth Shetty, Chandra Sekhar Seelamantula\\
    Indian Institute of Science\\
    Bengaluru, Karnataka, India\\
    {\tt\small \{saisuchithm, aditeyn, rohit1, nishanths, css\}@iisc.ac.in}
}

\begin{document}
    \twocolumn[{
    \maketitle
    \begin{center}
        \centering
        \captionsetup{type=figure}
        \includegraphics[width=\textwidth]{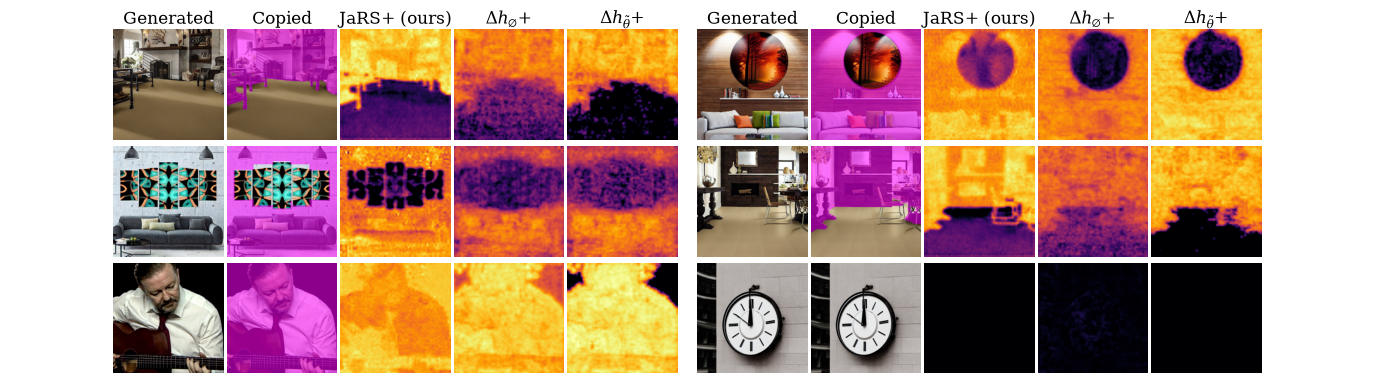}
        \captionof{figure}{\textbf{Qualitative comparison of maps for localized memorization on Stable Diffusion v1.4} for randomly sampled test instances. We present four partially memorized images (row 1--2), one whole-image memorized (bottom left), and one without any memorized region (bottom right) which should return an empty mask. Every method is shown in its two-stage form of \cref{tab:main}, which returns an empty map when its Stage 1 does not detect memorization. All maps are displayed after the $\log(1+\cdot)$ transformation.}
        \label{fig:teaser}
    \end{center}%
    }]
    \begin{abstract}
    Diffusion models sometimes memorize and reproduce regions of their training images. Addressing this problem is critical because unintended memorization creates legal liabilities, such as copyright infringement, and poses major privacy risks by leaking sensitive data. We aim to localize these regions.
    Existing contrast-based methods use the diagonal of the denoiser Jacobian, which reflects the posterior variance of each pixel. However, a small diagonal entry is ambiguous: it arises both from memorization and from averaging over many neighboring pixels in smooth regions. To resolve this ambiguity, these methods subtract a reference map, which requires a second evaluation with a null prompt or an earlier checkpoint. Unconditional models have no null prompt, and earlier checkpoints are rarely released.
    We show that Jacobian \emph{row sums} (JaRS), unlike the diagonal, separate memorization from local averaging without a reference map. Pairing JaRS with an image-level detection score computed from the JaRS map at an earlier denoising step yields JaRS+.
    Across Stable Diffusion v1.4 and v2.1, Realistic Vision v5.1, and an unconditional diffusion Transformer (DiT), JaRS+ outperforms contrast-based localization methods in 8 of 10 settings. 
\end{abstract}

    \section{Introduction} \label{sec:intro}

Diffusion models~\cite{sohl-dickstein-diffusion-models, ho-diffusion-models, song2021scorebased, song2021denoising} have achieved remarkable success in image synthesis, yet they are known to memorize and replicate portions of their training data~\cite{carlini2023, somepalli2023}. While certain prompts cause text-to-image models to regenerate an entire training image almost verbatim, a common and subtler form of replication is \emph{local memorization}: the model reproduces a specific region of a training image such as a foreground product, an interior layout, or a watermark---while synthesizing novel content in the rest of the image~\cite{somepalli2023understanding, webster2023reproducible}. Such partial copies raise significant copyright and privacy concerns. Because the memorized region is embedded within newly generated content, localizing the exact pixels that are copied tells us what the model has memorized and makes it possible to selectively remove the memorized region while preserving the surrounding generation.

Recently, \citet{kimlee2026} showed that memorized regions can be localized directly from the local geometry of the denoiser. By Tweedie's formula~\cite{robbins1992empirical, miyasawa1961empirical, efron2011tweedie}, the Jacobian $J$ of the denoiser with respect to its noisy input is proportional to the posterior covariance of the clean image, and is directly related to the Hessian of the log density of the noisy marginal distribution. When a diffusion model memorizes a region, its posterior distribution over that region collapses~\cite{jeon2025sail, kimlee2026}; the posterior variance and hence the Jacobian become small, so the denoiser largely ignores perturbations to its noisy input. \citet{kimlee2026} measure this collapse by using randomized Hutchinson probes~\cite{hutchinson1989stochastic, bekas2007estimator} to estimate the diagonal entries of the denoiser Jacobian (equivalently, the diagonal of the negative Hessian of the log-density). As \citet{kimlee2026} note, however, a small diagonal entry $J_{ii}$ is ambiguous on its own. Even a model that does not memorize has a small $J_{ii}$ in smooth regions such as a uniform background. There, the denoiser estimates each pixel by averaging its $N$ noisy neighbors, so pixel $i$'s own weight $J_{ii}$ is only about $1/N$. To remove this data-driven curvature~\cite{ross2025geometric}, \citet{kimlee2026} rely on a \emph{contrast}: they subtract a reference diagonal map computed either with an unconditional null prompt ($\Delta h_\varnothing$) or with an earlier, less-trained checkpoint of the same model ($\Delta h_{\tilde\theta}$).

While contrast-based localization is effective on standard text-to-image models, requiring a second reference evaluation introduces fundamental limitation: a null prompt exists only for conditional models trained with classifier-free guidance~\cite{ho2022cfg}, and is absent in unconditional models. Similarly, model developers rarely release intermediate training checkpoints alongside final weights. In this paper, we ask whether memorized regions can be separated from data-driven smoothing directly from the target model alone, without any reference prompt or secondary checkpoint. Our key insight is that while the diagonal entry $J_{ii}$ cannot distinguish between a row whose weights are spread across $N$ neighboring pixels and a row whose entries vanish altogether, the \emph{row sum} $\sum_j J_{ij}$ separates the two. 

Building on this principle, we introduce \textbf{JaRS} (\textbf{Ja}cobian \textbf{R}ow \textbf{S}ums), a contrast-free localization map that measures how far the per-channel row sums of the denoiser Jacobian fall below $1$. Because the row sum is a linear functional of the Jacobian, JaRS requires no randomized probes: for a latent space with $C$ channels, the map is obtained from $C$ deterministic Jacobian-vector products (JVPs). Furthermore, when memorized and unmemorized generations are evaluated together, a localizer must also output an empty mask on unmemorized images. To this end, we propose \textbf{JaRS+}, a two-stage procedure: it first uses an early-step summary statistic of the JaRS map to decide whether an image contains memorized content, and returns the pixel-level JaRS map only if it does. Our main contributions are summarized as follows:
\begin{itemize}
    \item We show, in a simple analytical model and empirically on full Jacobians of trained denoisers, that the row sums of the denoiser Jacobian concentrate near $1$ on generated regions and drop on memorized regions.
    \item We propose JaRS and its two-stage extension JaRS+. Evaluated under the balanced protocol of \citet{kimlee2026} across Stable Diffusion v1.4, Stable Diffusion v2.1, Realistic Vision v5.1, and an unconditional DiT-B/2 model, JaRS+ achieves the best localization mIoU in eight of ten settings. Over the strongest baseline, it improves mIoU by up to $6.9$ points on Stable Diffusion and from $0.872$ to $0.928$ on the DiT. In prompt- and sample-level detection, JaRS+ is competitive with contrast-based detectors.
\end{itemize}

    \section{Related Work} \label{sec:related}

Early studies~\cite{carlini2023,somepalli2023,webster2023reproducible} demonstrated that diffusion models can memorize and regenerate training images. \citet{carlini2023} extracted near-exact copies of training images from Stable Diffusion and unconditional DDPMs~\cite{ho-diffusion-models} using $\ell_{2}$ distance, while \citet{somepalli2023} used feature space metrics such as Self Supervised Copy Detection (SSCD)~\cite{pizzi2022self}. \citet{webster2023reproducible} curated a benchmark of memorized prompts in Stable Diffusion and categorized them into three groups: \emph{matching verbatims} (MV), which reproduce a training image paired with its training caption; \emph{retrieved verbatims} (RV), where a prompt triggers a verbatim copy of a training image associated with a different caption; and \emph{template verbatims} (TV), where the model reproduces a fixed spatial region (template) of a training image while generating diverse variations in the remaining regions across random seeds. 

Subsequent works investigated the causes of memorization and proposed inference time detection and mitigation strategies. \citet{somepalli2023understanding} linked replication to data duplication and text conditioning, proposing random token addition or replacement (RTA) to perturb memorized prompts. \citet{wen2024detecting} observed that memorized prompts induce abnormally large text-conditional noise predictions and mitigated memorization by optimizing the prompt embedding at inference time, \citet{chen2025enhancing} re-anchored and searched prompts within classifier-free guidance to improve the privacy-utility trade-off, and \citet{jeon2025sail} optimized the initial Gaussian latent to steer trajectories away from sharp probability modes with Sharpness-Aware Initialization for Latent Diffusion (SAIL). These methods modify the prompt or the initial latent of the whole generation.

Only recently has attention turned to pixel-level localization of memorized content. \citet{chen2025bright} argued that memorization should be analyzed locally and not only for whole images. They observed that at the final denoising step the cross-attention of the end-of-text token puts abnormally high weight on specific image patches (“Bright Ending”), and they used the resulting attention map to identify locally memorized regions. They also used this map to weight the noise-prediction magnitude of \citet{wen2024detecting}, both as the detection signal and as the loss for prompt-embedding optimization, so that detection and mitigation focus on the locally memorized region.

Without relying on cross-attention, \citet{kimlee2026} interpret memorization as a collapse of the posterior $p(\xo \mid \xt)$~\cite{jeon2025sail} and estimate the diagonal of the negative Hessian of $\log p_t(\xt)$ at the final denoising step via Hutchinson-type probing~\cite{hutchinson1989stochastic, bekas2007estimator}. Because a small diagonal also arises in smooth regions of ordinary images, where a generalizing denoiser averages over a wide neighborhood, \citet{kimlee2026} subtracted a reference diagonal computed under the null prompt or an earlier training checkpoint, and established the localization benchmark that we adopt in our experiments.

Concurrent work by \citet{gu2026tacotrajectoryawarecurvatureobservation} (TACO) further refines the null-prompt diagonal contrast of \citet{kimlee2026} by aggregating and spatially smoothing late-stage curvature differences to guide localized mitigation. While we share the posterior-collapse perspective, our work departs from both \citet{kimlee2026} and \citet{gu2026tacotrajectoryawarecurvatureobservation} by examining the \emph{row sums} of the denoiser Jacobian rather than its diagonal---which eliminates the need for any contrast prompt or reference checkpoint (see App.~\ref{supp:taco} for a discussion on TACO).

Tweedie's formula~\cite{robbins1992empirical, miyasawa1961empirical, efron2011tweedie} connects the optimal denoiser to the score function $\nabla \log p_t(\xt)$, and differentiating it once more links the Jacobian of the denoiser to the posterior covariance $\Cov[\xo \mid \xt]$. In classical and deep image denoising~\cite{milanfar2012tour, romano2017little, mohan2020robust, kadkhodaie2024generalization}, the Jacobian of a denoiser is interpreted as an input-dependent, locally adaptive linear filter; in particular, mean-preserving smoothing filters are \emph{row-stochastic}, meaning that their row weights sum to $1$~\cite{milanfar2012tour}. In diffusion posterior sampling for inverse problems, \citet{boys2024tweedie} multiplied the denoiser Jacobian by an all-ones vector as a computational shortcut to approximate the diagonal of $\Cov[\xo \mid \xt]$. \citet{jaziri2026spectral} studies the global leading singular value of the denoiser Jacobian along trajectories to characterize whole-image memorization basins.

    \section{Background and Preliminaries} \label{sec:setup}

We establish diffusion notation, link the denoiser Jacobian to the posterior, and review framework of \citet{kimlee2026}.

\subsection{Setup} \label{sec:setup:jacobian}

Let $\xo \in \R^n$ denote a clean image (or the latent representation of an image) drawn from the data distribution $p_0$. A forward diffusion process progressively corrupts $\xo$ with Gaussian noise over time $t \in [0, 1]$ according to
\begin{align} \label{eq:forward}
    \xt = \alpha_t \xo + \sigma_t \x{\epsilon}, \qquad \x{\epsilon} \sim \mathcal{N}(0, I)
\end{align}
where $\alpha_t, \sigma_t > 0$ define the noise schedule. Standard variance-preserving diffusion models such as Stable Diffusion~\cite{rombach2022high, ho-diffusion-models, song2021denoising} choose $\alpha_t^2 + \sigma_t^2 = 1$ and train a neural network $\x{\epsilon}_{\x{\theta}}(\xt, t)$ to predict the injected noise $\x{\epsilon}$. Flow-matching models~\cite{lipman2023flow, peebles2023scalable} instead use the linear interpolation schedule $\alpha_t = 1 - t$ and $\sigma_t = t$, training a velocity network $\x{v}_{\x{\theta}}(\xt, t)$ to predict $\x{\epsilon} - \xo$. In either formulation, given a noisy input $\xt$, the network produces a one-step estimate of the clean data point:
\begin{align} \label{eq:xhat}
    \xhat(\xt) = \frac{\xt - \sigma_t \x{\epsilon}_{\x{\theta}}(\xt, t)}{\alpha_t}
\end{align}
Let $p_t(\xt)$ denote the marginal density of $\xt$ induced by \cref{eq:forward}. By Tweedie's formula~\cite{robbins1992empirical, miyasawa1961empirical, efron2011tweedie}, the optimal noise predictor $\x{\epsilon}^{\star}(\xt, t)$ is
\begin{align*}
    \x{\epsilon}^{\star}(\xt, t) = -\sigma_t \nabla_{\xt} \log p_t(\xt)
\end{align*}
and the corresponding optimal denoiser is given by
\begin{align} \label{eq:tweedie}
    \E[\xo \mid \xt] = \frac{\xt + \sigma_t^2 \nabla_{\xt} \log p_t(\xt)}{\alpha_t}
\end{align}
To study the first-order sensitivity of the denoiser to perturbations of its noisy input $\xt$, we work with the scaled denoiser $D(\xt) = \alpha_t \xhat(\xt)$. Differentiating $D(\xt)$ with respect to $\xt$ yields the \emph{denoiser Jacobian} $J(\xt) \in \R^{n \times n}$:
\begin{align}
    J(\xt) = \frac{\partial D(\xt)}{\partial \xt} = I - \sigma_t \frac{\partial \x{\epsilon}_{\x{\theta}}(\xt, t)}{\partial \xt}.
    \label{eq:jac}
\end{align}
When $\x{\epsilon}_{\x{\theta}}$ equals the optimal noise predictor $\x{\epsilon}^{\star}$, substituting \cref{eq:tweedie} into \cref{eq:jac} and differentiating a second time via Tweedie's second-order identity gives
\begin{align} \label{eq:cov}
    J(\xt) = I + \sigma_t^2 H(\xt) = \frac{\alpha_t^2}{\sigma_t^2} \Cov[\xo \mid \xt]
\end{align}
where $H(\xt) = \nabla_{\xt}^2 \log p_t(\xt)$ is the Hessian of the log-density of the noisy marginal $p_t$. Additionally, one can view the Jacobian as an adaptive linear filter~\citep{mohan2020robust}. Locally around $\xt$, the $i$-th row of the Jacobian, $J_{i,:} = \frac{\partial D_i(\xt)}{\partial \xt}$, specifies how the denoised estimate at coordinate $i$ responds to infinitesimal changes at input coordinate $j \in \{1,\dots,n\}$. For a bias-free network, the denoiser is exactly a locally adaptive linear filter, $D(\xt) = J(\xt)\,\xt$, whose $i$-th row $J_{i,:}$ contains the spatial averaging weights used to reconstruct pixel $i$ from its noisy neighborhood~\cite{mohan2020robust, kadkhodaie2024generalization}. For a general network, the rows still measure local sensitivity: $\sum_j J_{ij}$ is the response of $D_i$ to a uniform shift of its input. We verify in \cref{sec:exp:rows} that, at intermediate noise, trained diffusion denoisers behave as mean-preserving filters on non-memorized content.

\subsection{Curvature-Based Localization} \label{sec:setup:confound}
We formalize the relationship between the Jacobian of the denoiser and memorization. When a diffusion model memorizes a training image or a spatial region $R$ of it, the posterior $p(\xo \mid \xt)$ over $R$ collapses~\cite{jeon2025sail}, so the posterior variance $\Var(x_{0,i} \mid \xt)$ vanishes for every memorized pixel $i \in R$. By \cref{eq:cov}, this implies $J_{ii}(\xt) \approx 0$ meaning the denoised output $D_i(\xt)$ stays fixed to the training content under small perturbations of $\xt$. \citet{kimlee2026} turn this observation into a localization metric by examining the diagonal of the negative log-density Hessian $-H(\xt)$, which by \cref{eq:cov} is
\begin{align} \label{eq:diagh}
    \sigma_{t}\diag(-H(\xt))_i = \frac{1 - J_{ii}(\xt)}{\sigma_t}
\end{align}
\citet{kimlee2026} estimate $\sigma_{t}\diag(-H(\xt))$ at the final denoising step (we follow their official implementation), marking pixels with large values of $\sigma_{t}\diag(-H(\xt))$ (equivalently, small values of $J_{ii}$) as memorized. However, a small diagonal entry $J_{ii}(\xt)$ does \emph{not} identify memorization. To see why, consider the following example.

\begin{example} \label{ex:confound}
    Suppose a denoiser reconstructs a smooth image region (such as a uniform wall or clear sky) of $N$ pixels by taking an unweighted average of the $N$ noisy pixels in that region $D_i(\xt) = \frac{1}{N}\sum_{j \in R} x_{t,j}$ where $i \in R$. The Jacobian is given by $J_{ij}(\xt) = 1 / N$ for all $j \in R$. Consequently, the diagonal is $J_{ii}(\xt) = 1 / N$, which approaches $0$ as the averaging neighborhood $N$ grows large. Thus, on a smooth region of size $N \gg 1$, the diagonal entry $J_{ii} = 1/N$ is nearly indistinguishable from the diagonal entry $J_{ii} = 0$ of a memorized region where the denoiser outputs a stored constant $D_i(\xt) = c_i$.
\end{example}

Following \citet{ross2025geometric}, \citet{kimlee2026} refer to the large values of $\diag(-H)_i$ as \emph{data-driven curvature}, in contrast to \emph{memorization-driven curvature}. As discussed in \cref{ex:confound}, raw diagonal (\cref{eq:diagh}) cannot differentiate between these two. To cancel the data-driven component, \citet{kimlee2026} use a \emph{contrast}, which assumes that a reference evaluation ($\varnothing$ or $\tilde{\x{\theta}}$) exhibits the same data-driven smoothing on generic regions, so that subtracting it cancels the background response and isolates the memorized region. For a conditional model $\x{\epsilon}_{\x{\theta}}(\xt, t, y)$ with prompt $y$, they subtract the diagonal curvature computed either with null prompt $\varnothing$,
\begin{align} \label{eq:dh_null}
    (\Delta h_\varnothing)_i = \diag(-H_{\x{\theta}}(\xt, t, y))_i - \diag(-H_{\x{\theta}}(\xt, t, \varnothing))_i
\end{align}
or with a second, less-trained checkpoint $\tilde{\x{\theta}}$,
\begin{align} \label{eq:dh_ckpt}
    (\Delta h_{\tilde\theta})_i = \diag(-H_{\x{\theta}}(\xt, t, y))_i - \diag(-H_{\tilde{\x{\theta}}}(\xt, t, y))_i
\end{align}
\citet{kimlee2026} also evaluate two \emph{score contrasts}~\cite{wen2024detecting} given by
\begin{align} \label{eq:ds}
    (\Delta s_\varnothing)_i &= \big(\x{s}_{\x{\theta}}(\xt, t, y) - \x{s}_{\x{\theta}}(\xt, t, \varnothing)\big)_i^2, \nonumber \\
    (\Delta s_{\tilde\theta})_i &= \big(\x{s}_{\x{\theta}}(\xt, t, y) - \x{s}_{\tilde{\x{\theta}}}(\xt, t, y)\big)_i^2
\end{align}
where $\x{s}_{\x{\theta}} = -\x{\epsilon}_{\x{\theta}} / \sigma_t$. Following the implementation of \citet{kimlee2026}, a 2D spatial map is formed by summing the coordinate-wise values across the $C$ channels at each latent pixel.

As noted in \cref{sec:intro}, however, computing \cref{eq:dh_null}, \cref{eq:dh_ckpt}, or \cref{eq:ds} requires a reference evaluation that is unavailable for unconditional models or when suitable pre-memorization checkpoints are not released. This motivates a central question: \emph{Can we distinguish local smoothing from memorization using only the target model itself?} \cref{ex:confound} suggests a route: while the diagonal entry $J_{ii}$ shrinks to $1/N$ under local averaging, the \emph{sum of the entries across the row} is $\sum_{j \in R} J_{ij} = N \cdot (1/N) = 1$, whereas under memorization ($J_{ij} = 0$) the row sum vanishes. We formalize this observation in the next section.

    \section{Jacobian Row Sums (JaRS)} \label{sec:method}

We now present our contrast-free approach to localizing memorized regions. In \cref{sec:method:jrs}, we define the \textbf{JaRS} (\textbf{Ja}cobian \textbf{R}ow \textbf{S}ums) localization map. In \cref{sec:method:compute}, we show how JaRS can be computed using Jacobian-vector products (we verify the predictions of this section on full denoiser Jacobians in \cref{sec:exp:rows}). Finally, we introduce our two-stage detection-and-localization pipeline \textbf{JaRS+} in \cref{sec:method:jrsplus}.

\begin{figure*}
    \centering
    \includegraphics[width=\textwidth]{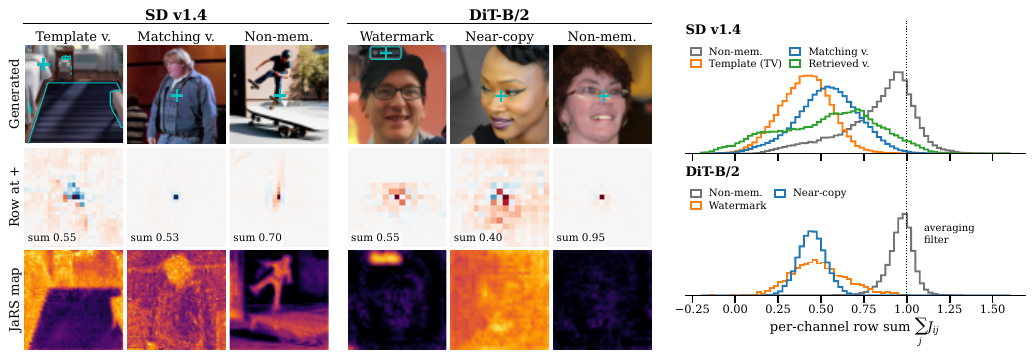}
    \caption{\textbf{Full denoiser Jacobian rows of Stable Diffusion v1.4 (step $15$) and DiT-B/2 (step $35$).} \emph{Left:} For one development image from each category, we show the generated image (copied region outlined in cyan), the row $J_{i,:}$ of the full Jacobian at the marked pixel `$+$' together with its sum, and the JaRS map \eqref{eq:score}. \emph{Right:} Empirical distributions, over development images, of the per-channel row sum $\sum_j J_{ij}$ across all pixels of non-memorized images and whole-image copies, and inside the copied region of template verbatims and watermarked samples. The vertical dotted line at $1$ marks the row sum of a mean-preserving averaging filter.}
    \label{fig:rows}
\end{figure*}

\subsection{Contrast-Free Localization} \label{sec:method:jrs}

Consider $\xt \in \R^n$ with $n = C \times H \times W$, where we index coordinates by $(c, i)$ for channel $c \in \{1, \dots, C\}$ and spatial pixel location $i \in \{1, \dots, HW\}$. Let $J_{(c,i),(c',j)}(\xt)$ denote the entry of the scaled denoiser Jacobian $J(\xt)$ in \cref{eq:jac}. Let $\one_c \in \R^n$ denote the channel indicator vector.

\begin{definition} \label{def:jars}
    At noise level $t$, the \emph{Jacobian Row Sum (JaRS)} score at spatial location $i \in \{1, \dots, HW\}$ is defined as
    \begin{align} \label{eq:score}
        s_t(i) &= \sum_{c=1}^{C} \left(1 - \sum_{j=1}^{HW} J_{(c,i),(c,j)}(\xt)\right) \nonumber \\
        &= C - \sum_{c=1}^{C} \left(J(\xt)\one_c\right)_{(c,i)}
    \end{align}
\end{definition}

$s_t(i)$ measures how far per-channel row sums fall below $1$. From \cref{eq:cov}, each channel term can be written equivalently as
\begin{align*}
    1 - \left(J(\xt)\one_c\right)_{(c,i)} = -\sigma_t^2 \left(H(\xt)\one_c\right)_{(c,i)}
\end{align*}
Thus, whereas \citet{kimlee2026} take the \emph{diagonal} of the negative log-density Hessian, JaRS takes its \emph{row sums}, and the constant $1$ per channel can be seen as a replacement for the contrast. By this identity, the row sum equals $1$ when $H(\xt)\one_c = 0$, i.e., when $\log p_t$ is flat along $\one_c$: slightly brightening or darkening a generic region yields an equally plausible image. 

As discussed in \cref{sec:setup:jacobian} and \cref{ex:confound}, the diagonal shrinks as $1/N$ for an $N$-pixel average, whereas the row sum does not decrease as the averaging neighborhood widens. Therefore, on generated regions, $(J(\xt)\one_c)_{(c,i)}$ is close to $1$ and the JaRS score $s_t(i)$ is close to $0$. Conversely, when the posterior collapses onto a stored image $\xo^*$, $p_t$ is locally close to $\mathcal{N}(\alpha_t \xo^*, \sigma_t^2 I)$, so $H(\xt) \approx -I/\sigma_t^2$: the denoiser ignores perturbations of $\xt$, $(J(\xt)\one_c)_{(c,i)}$ drops toward $0$, and $s_t(i)$ rises toward $C$.

The following proposition makes this precise in a simple analytical model of a homogeneous image region.

\begin{proposition} \label{prop:smooth}
    Consider a single-channel region $R$ of $N$ pixels whose clean values share a common intensity level, $x_{0,j} = c$ for all $j \in R$, under a Gaussian prior $c \sim \mathcal{N}(\mu, \tau^2)$ where $c$ is independent of $x_{0,k}$ for all $k \notin R$. Given a noisy observation $\xt$ following \cref{eq:forward}, the optimal denoiser Jacobian satisfies, for every pixel $i \in R$ and $j \in R$,
    \begin{align} \label{eq:prop_g}
        J_{ij}(\xt) = \frac{g}{N},
        \qquad
        g = \frac{N \alpha_t^2 \tau^2}{\sigma_t^2 + N \alpha_t^2 \tau^2} \in [0, 1)
    \end{align}
    Consequently, the diagonal entry at pixel $i \in R$ is $J_{ii}(\xt) = g/N$, whereas the row sum over $R$ is $\sum_{j \in R} J_{ij}(\xt) = g$.
\end{proposition}

\cref{prop:smooth} illustrates two properties of the JaRS map: robustness to neighborhood size and dependence on noise schedule. First, regarding neighborhood size $N$, the prior variance $\tau^2$ quantifies the model's uncertainty over $R$ prior to observing $\xt$. For generic, unmemorized regions where $\tau^2$ is large, the row sum $g$ approaches $1$ regardless of $N$ once $N \alpha_t^2 \tau^2 \gg \sigma_t^2$, while the diagonal entry $J_{ii}$ scales as $\mathcal{O}(1/N)$. In contrast, for memorized regions, the prior collapses onto a stored value ($\tau^2 \to 0$), yielding $g \to 0$ and driving the JaRS score $1 - g \to 1$. 

Second, whereas \citet{kimlee2026} evaluate curvature differences strictly at the final denoising step, \cref{eq:prop_g} identifies the noise-to-signal ratio $\sigma_t / \alpha_t$ as the primary control knob. In the high-noise regime ($\sigma_t^2 \gg N \alpha_t^2 \tau^2$), $g \approx 0$ even for unmemorized regions because $\xt$ is nearly uninformative of $\xo$. At near-zero noise ($\sigma_t^2 \to 0$), $g \to 1$ everywhere, unless the memorized content is stored with effective variance smaller than $\sigma_t^2 / (N \alpha_t^2)$. Between these extremes lies a broad intermediate noise window where $g \approx 1$ on generated regions and $g \ll 1$ on memorized regions; identifying and operating in this intermediate noise regime is a key step that takes us from final-step evaluation to JaRS and JaRS+.

\subsection{Computation} \label{sec:method:compute}
A practical advantage of \cref{def:jars} is that $s_t(i)$ is a linear functional of the Jacobian $J(\xt)$. Unlike the diagonal $\diag(J(\xt))$, which requires randomized Hutchinson probes~\cite{hutchinson1989stochastic, bekas2007estimator} to estimate, the row sums are obtained from the $C$ matrix-vector products $J(\xt)\one_1, \dots, J(\xt)\one_C$---one for each latent channel. Using \cref{eq:jac}, each product takes the form $J(\xt)\one_c = \one_c - \sigma_t J_{\x{\epsilon}}(\xt)\one_c$, where $J_{\x{\epsilon}}(\xt) = \frac{\partial \x{\epsilon}_{\x{\theta}}(\xt, t, y)}{\partial \xt}$ is the Jacobian of the noise predictor (with conditioning $y$ for conditional models, or omitted for unconditional models). We approximate each directional derivative $J_{\x{\epsilon}}(\xt)\one_c$ via a central finite difference:

\begin{align} \label{eq:fd}
    J_{\x{\epsilon}}(\xt)\one_c \approx \frac{\x{\epsilon}_{\x{\theta}}(\xt + h\one_c, t, y)  - \x{\epsilon}_{\x{\theta}}(\xt - h\one_c, t, y)}{2h}
\end{align}

For all our experiments, we used $h = 3 \times 10^{-3}$. Observe that computing the full JaRS map at a given timestep therefore requires $2C$ forward network evaluations, which can be run in a single batched forward pass without any backward pass or automatic differentiation graph (App.~\ref{supp:cost}).

\subsection{JaRS+: Two-Stage Detection and Localization} \label{sec:method:jrsplus}

The interpretation following \cref{prop:smooth} shows that the row-sum map should be evaluated at intermediate time steps. In realistic settings that mix memorized and non-memorized generations, however, a localizer must also output an empty mask on images that contain no copied content. As seen in the left tail of the non-memorized histogram for Stable Diffusion v1.4 in \cref{fig:rows}, backgrounds of ordinary images that the denoiser predicts confidently, such as uniform studio backgrounds, also exhibit low posterior variance at intermediate noise levels. If a single pixel threshold is applied across all images at step $\ell$ without an image-level check, these confident backgrounds produce false-positive masks on non-memorized images (compare the JaRS and JaRS+ rows of \cref{tab:main}).

Analyzing the JaRS map across the denoising trajectory shows that \emph{image-level detection} and \emph{pixel-level localization} work on two different timescales. In conditional models, the prompt locks in memorized content during the earliest, high-noise denoising steps. At these steps, the JaRS map is already elevated across the whole image on memorized trajectories, while ordinary trajectories are still dominated by the initial Gaussian noise. At intermediate noise, the spatial boundaries between copied templates and generated backgrounds become sharp. On the unconditional models, image-level separation appears later (App.~\ref{supp:traj}, \cref{fig:supp:traj}). Exploiting these two timescales takes us from JaRS to JaRS+, a two-stage procedure that pairs an image-level detection stage at an step $d$ with a pixel-level localization stage at step $\ell$:
\begin{enumerate}
    \item \textbf{Stage 1 (Image-level detection):} At denoising step $d$, we compute the upsampled JaRS map $s_{t_d}$ and summarize it by a scalar statistic $z_d = \phi(s_{t_d})$ over its pixels, where $\phi$ is the mean, the median $Q_{0.5}$, or the percentile $Q_{0.9}$ or $Q_{0.99}$. The generation is flagged as containing memorized content if and only if $z_d > \tau$, where $\tau$ is a scalar detection threshold.
    \item \textbf{Stage 2 (Pixel-level localization):} If $z_d \le \tau$, the image is declared unmemorized and JaRS+ outputs an identically zero map. If $z_d > \tau$, JaRS+ returns the JaRS localization map $s_{t_\ell}$ evaluated at step $\ell$.
\end{enumerate}
We now describe how we select $\ell$, $d$, $\phi$, and $\tau$. For each model, these hyperparameters are selected on a held-out development split.
\begin{itemize}
    \item \textbf{Localization step $\ell$:} Chosen from the candidate steps $\{5, 10, \dots, 45, 49\}$ of the $50$-step sampler to maximize the JaRS mean IoU on the memorized development images.
    \item \textbf{Detection step $d$ and statistic $\phi$:} Chosen from $\{0, 5, \dots, 45, 49\}$ and the four statistics above to maximize the image-level detection AUROC of $z_d$ between memorized and non-memorized images on the balanced development split of the model's broadest evaluation setting (\cref{sec:exp:setup})
    \item \textbf{Detection threshold $\tau$:} Chosen from candidate quantiles ($5\%, 7.5\%, \dots, 95\%$) of $z_d$ on the same balanced development split to maximize the resulting two-stage development mIoU. When labeled memorized examples are not available to tune $\tau$, $\tau$ can instead be set directly to the $95$th percentile of $z_d$ on ordinary non-memorized images ($5\%$ development false-positive rate), which we show in \cref{supp:detect} stays within $0.033$ mIoU of the tuned threshold on all four models.
\end{itemize}
All ties above are broken in favor of the earlier step. Selection of $\ell, d, \tau$ above is discussed for $50$ sampling steps, but can be extended easily to arbitrary number of sampling steps.

    \section{Experiments} \label{sec:exp}

We evaluate JaRS and JaRS+ across text-to-image diffusion models and unconditional Diffusion Transformer (DiT). We first describe our benchmarks, evaluation protocol, and baselines (\cref{sec:exp:setup}), verify the behavior of row sums predicted in \cref{sec:method:jrs} on full denoiser Jacobians (\cref{sec:exp:rows}), and then present our main pixel-level localization results (\cref{sec:exp:main}). Finally, we evaluate prompt-level memorization detection (\cref{sec:exp:promptdet}). In App.~\ref{supp:mitig}, we further show that restricting the inference-time mitigation of \citet{wen2024detecting} to the region localized by JaRS lowers the copy rate from $20.5\%$ to $9.0\%$ and the perceptual change outside the region from $0.708$ to $0.158$ LPIPS.

\begin{table*}
    \centering
    \small
    \caption{\textbf{Pixel-level localization of memorized regions on the test splits (mIoU)} under the balanced protocol of \citet{kimlee2026}. Baselines of \citet{kimlee2026} is applied in the two-stage form as JaRS+ (`+'), and uses $16$ Hutchinson probes; checkpoint references for $\Delta h_{\tilde\theta}$ and $\Delta s_{\tilde\theta}$ are SD v1.1 (for SD v1.4 and Realistic Vision), SD v2.0-base (for SD v2.1), and the $5$k-step checkpoint, which is best on the DiT development split for both contrasts. JaRS denotes the map at step $\ell$ without Stage 1 of JaRS+. In settings with non-memorized images, an mIoU of $0.500$ corresponds to predicting an empty mask everywhere.}
    \setlength{\tabcolsep}{5pt}
    \begin{tabular}{@{}llcccccccccc@{}}
        \toprule
        & & \multicolumn{3}{c}{SD v1.4} & \multicolumn{2}{c}{SD v2.1} & \multicolumn{2}{c}{Realistic Vision} & \multicolumn{3}{c}{DiT (unconditional)} \\
        \cmidrule(lr){3-5} \cmidrule(lr){6-7} \cmidrule(lr){8-9} \cmidrule(l){10-12}
        Method & Reference & TV & TV+N & All & TV & TV+N & TV & All & WM & WM+N & All \\
        \midrule
        Raw diagonal+~\cite{kimlee2026} & -- & 0.768 & 0.877 & 0.909 & 0.811 & 0.896 & 0.844 & 0.914 & 0.038 & 0.500 & 0.662 \\
        $\Delta h_\varnothing$+~\cite{kimlee2026} & null prompt & 0.900 & 0.950 & \textbf{0.945} & 0.943 & 0.967 & 0.930 & \textbf{0.951} & \multicolumn{3}{c}{\emph{not available}} \\
        $\Delta h_{\tilde\theta}$+~\cite{kimlee2026} & checkpoint & 0.879 & 0.936 & 0.926 & 0.918 & 0.956 & 0.895 & 0.927 & 0.790 & 0.870 & 0.872 \\
        $\Delta s_\varnothing$+~\cite{kimlee2026} & null prompt & 0.841 & 0.921 & 0.919 & 0.911 & 0.946 & 0.876 & 0.922 & \multicolumn{3}{c}{\emph{not available}} \\
        $\Delta s_{\tilde\theta}$+~\cite{kimlee2026} & checkpoint & 0.882 & 0.941 & 0.912 & 0.921 & 0.960 & 0.847 & 0.844 & 0.606 & 0.803 & 0.738 \\
        \midrule
        JaRS (ours) & -- & 0.969 & 0.619 & 0.637 & 0.980 & 0.571 & 0.949 & 0.612 & 0.836 & 0.873 & 0.909 \\
        \textbf{JaRS+ (ours)} & -- & \textbf{0.969} & \textbf{0.983} & 0.931 & \textbf{0.980} & \textbf{0.989} & \textbf{0.944} & 0.850 & \textbf{0.819} & \textbf{0.895} & \textbf{0.928} \\
        \bottomrule
    \end{tabular}
    \label{tab:main}
\end{table*}

\subsection{Benchmarks, Protocol, and Baselines} \label{sec:exp:setup}
We adopt the memorization localization benchmark of \citet{kimlee2026}, which is built on the memorized prompts identified by \citet{webster2023reproducible} across three settings: Stable Diffusion v1.4 (SD v1.4), Stable Diffusion v2.1-base (SD v2.1)~\cite{rombach2022high}, and Realistic Vision v5.1. The benchmark comprises three categories of generations: Template verbatims (TV), Matching and Retrieved verbatims (MV, RV) come under global memorization and non-memorized prompts (N). We generate all images and ground-truth masks using the official code released by \citet{kimlee2026}.

To evaluate localization in an unconditional setting where no null prompt exists and with exact ground-truth masks, we train an unconditional DiT (\textbf{DiT-B/2}~\cite{peebles2023scalable}) using flow matching~\cite{lipman2023flow} on the $70{,}000$ FFHQ~\cite{karras2019style} face images at $256 \times 256$ resolution. Prior to training, we inject two controlled forms of memorization into the dataset: \emph{local memorization (WM):} a fixed semi-transparent watermark is pasted at a random spatial location into $10\%$ of the training images and \emph{global memorization (near-copies):} $20$ clean training images are duplicated $100$ times each. Generated samples are labeled as watermarked (with the watermark footprint as the ground-truth mask), near-copies (all-ones mask), or non-memorized (empty mask), and we retain $300$ watermarked, $300$ non-memorized, and all $139$ near-copy samples (App.~\ref{supp:bench}).

We follow the balanced evaluation protocol of \citet[App.~F.2]{kimlee2026} across all settings reported in their work: \textbf{TV} (\textbf{WM} for DiT) evaluates localization exclusively on locally memorized images, \textbf{TV+N} (\textbf{WM+N}) adds non-memorized images, and \textbf{All} further adds globally memorized images, with each category subsampled once to the size of the smallest. Every localization map is clipped at $0$, transformed via $x \mapsto \log(1 + x)$, min-max normalized across the evaluated set, and binarized using the single threshold that maximizes the mean Intersection-over-Union (\textbf{mIoU}) across images.

We partition all prompts (along with their seeds and templates) and DiT samples into a $20\%$ development split (indices divisible by $5$) and $80\%$ test split. All hyperparameters of JaRS+ and of the baselines ($\ell$, $d$, $\phi$, and $\tau$; \cref{sec:method:jrsplus}) are selected on the development split, and all experimental results are reported on the test split. For JaRS+, the development split selects $(\ell, d) = (15, 5)$ for SD v1.4 and SD v2.1, $(35, 5)$ for Realistic Vision, and $(35, 25)$ for the DiT, with $\phi = Q_{0.99}$ on all models except SD v2.1 ($\phi = Q_{0.9}$).

We compare JaRS+ against the maps of \citet{kimlee2026}: the contrast-free \textit{raw diagonal}, the \textit{null-prompt contrast} $\Delta h_\varnothing$ (\cref{eq:dh_null}), the \textit{checkpoint contrast} $\Delta h_{\tilde\theta}$ (\cref{eq:dh_ckpt}), and the two \textit{score contrasts} $\Delta s_\varnothing$ and $\Delta s_{\tilde\theta}$ (\cref{eq:ds}). For a fair comparison, we apply the two-stage procedure of JaRS+ (\cref{sec:method:jrsplus}) to every baseline, denoted by `+': Stage 1 uses a statistic of the baseline's own map at step $d$, Stage 2 returns its map at step $\ell$, and $(\ell, d, \phi, \tau)$ are selected on the development split with the same candidates and rules as for JaRS+ (App.~\ref{supp:bench}). On the DiT, only the checkpoint contrasts are applicable, with the reference selected on the development split among five checkpoints saved during our training run. App.~\ref{supp:published} reports the baselines in their published configuration, at the final sampling step and without Stage 1.

\subsection{Empirical Verification on Full Jacobians} \label{sec:exp:rows}

To verify whether trained diffusion models follow the predicted behavior in \cref{sec:method:jrs}, we explicitly form the full $n \times n$ denoiser Jacobian $J(\xt)$ and inspect its rows directly (\cref{fig:rows}). On the DiT ($n = 4{,}096$), we materialize the Jacobian at $\ell = 35$ for $10$ randomly selected development samples from each category: watermarked, near-copies, and non-memorized. On non-memorized samples, each row $J_{i,:}$ is a localized smoothing kernel centered at pixel $i$, and the per-channel row sums concentrate sharply around $1$ ($80\%$ of values in $[0.87, 1.05]$ with $0.97$ as median), confirming that the denoiser acts as a mean preserving local filter. The unmemorized background of watermarked samples behaves identically (median is $0.98$). In contrast, inside the memorized watermark region, the row sums drop to a median of $0.48$, and on whole-image near-copies they fall to $0.45$. Across all pixels, the row sum separates watermark pixels from non-memorized pixels with an AUROC of $0.998$. 

The same phenomenon holds on SD v1.4 ($n = 16{,}384$), where we compute the full Jacobian at the localization step $\ell = 15$ for $37$ development images: $10$ TV, $10$ MV, all $7$ RV of the development split, and $10$ non-memorized images. The per-channel row sums have a median of $0.88$ on non-memorized images and $0.89$ outside the copied template, falling to $0.41$ inside the template region, $0.54$ on matching verbatims, and $0.58$ on retrieved verbatims. App.~\ref{supp:l1} repeats this analysis on SD v2.1 and also compares the row sum with the $\ell_1$ norm of the row.

\subsection{Memorization Localization Results} \label{sec:exp:main}

\begin{table}[t]
    \centering
    \small
    \caption{\textbf{Prompt- and sample-level memorization detection on the test split}, measured by AUROC (AUC) and true-positive rate at $1\%$ false-positive rate (TPR). Every method scores a prompt (a sample on the DiT) with its own Stage 1 statistic at its step $d$, as in \cref{tab:main}. For Stable Diffusion, we exclude memorized prompts that are in the development split of the localization benchmark (App.~\ref{supp:promptdet}).}
    \resizebox{\columnwidth}{!}{%
    \begin{tabular}{@{}lcccccc@{}}
        \toprule
        & \multicolumn{2}{c}{SD v1.4} & \multicolumn{2}{c}{SD v2.1} & \multicolumn{2}{c}{DiT} \\
        \cmidrule(lr){2-3} \cmidrule(lr){4-5} \cmidrule(l){6-7}
        Method & AUC & TPR & AUC & TPR & AUC & TPR \\
        \midrule
        Raw diagonal+~\cite{kimlee2026} & 0.982 & 0.938 & 0.995 & 0.943 & 0.844 & 0.308 \\
        $\Delta h_\varnothing$+~\cite{kimlee2026} & 0.963 & 0.913 & 0.985 & 0.957 & \multicolumn{2}{c}{\emph{not available}} \\
        $\Delta h_{\tilde\theta}$+~\cite{kimlee2026} & 0.960 & 0.866 & 0.968 & 0.921 & \textbf{1.000} & \textbf{1.000} \\
        $\Delta s_\varnothing$+~\cite{kimlee2026} & 0.963 & 0.930 & 0.979 & 0.950 & \multicolumn{2}{c}{\emph{not available}} \\
        $\Delta s_{\tilde\theta}$+~\cite{kimlee2026} & \textbf{0.987} & \textbf{0.947} & \textbf{1.000} & \textbf{1.000} & \textbf{1.000} & 0.997 \\
        \midrule
        \textbf{JaRS+ (ours)} & 0.960 & 0.877 & 0.990 & 0.964 & 0.998 & 0.951 \\
        \bottomrule
    \end{tabular}}
    \label{tab:promptdet}
\end{table}

\cref{tab:main} reports the quantitative localization results across all four models, and \cref{fig:teaser} visualizes representative localization maps on SD v1.4. On both SD v1.4 and v2.1, JaRS+ achieves the highest localization accuracy in the TV and TV+N settings despite using no reference prompt or checkpoint. Specifically, JaRS+ outperforms the strongest contrast baseline by $+6.9$ and $+3.3$ mIoU points on SD v1.4 and by $+3.7$ and $+2.2$ points on SD v2.1. In the All setting of SD v1.4, JaRS+ ($0.931$) trails $\Delta h_\varnothing$ ($0.945$). On TV, where every image is memorized, Stage 1 can only remove correct masks, so it lowers some baselines below their published configuration (e.g., $\Delta h_{\tilde\theta}$ from $0.913$ to $0.879$ on SD v1.4). Furthermore, JaRS+ is also performs the best on TV against the published configurations (App.~\ref{supp:published}).

On the unconditional DiT, where the null-prompt contrasts do not exist, JaRS+ outperforms the checkpoint contrast $\Delta h_{\tilde\theta}$ across all three settings, raising mIoU from $0.790$ to $0.819$ on WM, from $0.870$ to $0.895$ on WM+N, and from $0.872$ to $0.928$ on All. The checkpoint contrasts also depend on the choice of the reference checkpoint (App.~\ref{supp:ref}).

On Realistic Vision, JaRS+ is best on TV ($0.944$ vs.\ $0.930$ for $\Delta h_\varnothing$) but trails on All ($0.850$ vs.\ $0.951$). This gap arises from Stage 1 rather than the JaRS map: when Stage 1 is replaced by an oracle that passes exactly the memorized images, the same maps reach $0.945$ on All (App.~\ref{supp:detect}). Removing Stage 1 (JaRS in \cref{tab:main}) changes mIoU by at most $0.017$ on TV and WM, but lowers it in every setting with non-memorized images, sharply on the text-to-image models (e.g., from $0.983$ to $0.619$ on TV+N of SD v1.4).

It is instructive to compare JaRS+ with the \emph{raw diagonal}+ row of \cref{tab:main}. This row applies the same two-stage procedure, to the diagonal estimator of \citet{kimlee2026}, and like JaRS it uses no contrast. The two rows therefore differ mainly in using the row sum instead of the diagonal, and this choice raises TV mIoU from $0.768$ to $0.969$ on SD v1.4, from $0.811$ to $0.980$ on SD v2.1, and from $0.844$ to $0.944$ on Realistic Vision. The noise level matters as well. At the final step, the JaRS map is no better than marking the whole image as copied ($0.563$ and $0.648$ on TV of SD v1.4 and v2.1). In contrast, the step $\ell$ selected on the development split lies on a broad high-accuracy plateau at intermediate noise on all four models (App.~\ref{supp:traj}).

\subsection{Detecting Memorized Prompts and Samples} \label{sec:exp:promptdet}

Beyond pixel-level maps, the Stage 1 score $z_d$ of JaRS+ can directly serve as a detector of whether a prompt (or unconditional sample) triggers memorization. Following the prompt-detection protocol of \citet[Sec.~5.2]{kimlee2026}, each prompt is generated with four random seeds and scored by averaging the Stage 1 score over the seeds. However, for the DiT setting, individual samples are scored. Every method uses its own Stage 1 statistic and step $d$ from \cref{tab:main}. The positive and negative sets and the test split sizes are given in App.~\ref{supp:promptdet}.

As shown in \cref{tab:promptdet}, JaRS+ achieves $0.960$, $0.990$, and $0.998$ AUC across SD v1.4, SD v2.1, and the DiT without any reference evaluation. On SD v1.4, JaRS+ matches $\Delta h_{\tilde\theta}$+ ($0.960$) and comes within $0.003$ AUC of both null-prompt contrasts ($0.963$); on SD v2.1, it surpasses $\Delta h_\varnothing$+, $\Delta h_{\tilde\theta}$+, and $\Delta s_\varnothing$+ in both AUC and TPR, trailing only $\Delta s_{\tilde\theta}$+. On the unconditional DiT, JaRS+ approaches the checkpoint contrasts ($0.998$ vs.\ $1.000$ AUC) while outperforming the other contrast-free baseline, Raw diagonal+.

    \section{Conclusion} \label{sec:conclusion}

We have shown that memorized regions in diffusion models can be localized accurately with the help of row sums of the denoiser Jacobian (\textbf{JaRS}) without any contrast prompt or reference checkpoint. Combining JaRS with an image-level detection score at an earlier step yields JaRS+, which outperforms contrast-based localization in eight of ten settings and enables region-restricted mitigation (App.~\ref{supp:mitig}).

\textbf{Limitations and future work.} JaRS evaluates the denoiser at an intermediate noise level along a multi-step trajectory. Extending it to single-step generators, such as distilled models, remains an open direction. Further, JaRS+ selects $(\ell, d, \phi, \tau)$ on a small labeled development split, whereas the published configuration of \citet{kimlee2026} requires none on Stable Diffusion; $\tau$ can be calibrated from non-memorized images alone (App.~\ref{supp:detect}), but $\ell$ and $d$ still require memorized examples.

    \section*{LLM Usage Statement}

We used two large language model tools: Claude Opus 5.5 (Anthropic) and Gemini Argon (Google). The research question, the method, and the claims are the authors' own.

Opus 5.5 was used to set up the baselines, utility scripts for data processing and plotting. For editing, \LaTeX{} formatting and literature search, we used both Opus 5.5 and Gemini Argon. We used Gemini Argon to review the code and check for correctness.

The authors designed all experiments, including the benchmarks, the development/test splits, the evaluation protocol. The authors take full responsibility for all content of this submission.

    {
        \small
        \bibliographystyle{ieeenat_fullname}
        \bibliography{main}
    }
\clearpage
\appendix

\section{Proof of Proposition 1}
\begin{proof}
    Using \cref{eq:forward}, conditioned on $c$, the noisy observations in $R$ are independent with $x_{t,j} \mid c \sim \mathcal{N}(\alpha_t c, \sigma_t^2)$ for each $j \in R$. Using Eq. (2.116) of \citet{bishop2007}, the posterior distribution of $c \mid \xt$ is Gaussian with precision and mean given by    
    \begin{align*}
        \Lambda &= \frac{1}{\tau^2} + \frac{N \alpha_t^2}{\sigma_t^2}, \\
        \E[c \mid \xt] &= \Lambda^{-1} \left(\frac{\mu}{\tau^2} + \frac{\alpha_t}{\sigma_t^2} \sum_{j \in R} x_{t,j}\right) \\
                       &= \frac{\sigma_t^2 \mu + \alpha_t \tau^2 \sum_{j \in R} x_{t,j}}{\sigma_t^2 + N \alpha_t^2 \tau^2}.
    \end{align*}
    Since $x_{0,i} = c$ for every $i \in R$, the scaled optimal denoiser is $D_i(\xt) = \alpha_t \E[x_{0,i} \mid \xt] = \alpha_t \E[c \mid \xt]$. Differentiating $D_i(\xt)$ with respect to $x_{t,j}$ for any $j \in R$ yields
    \begin{align*}
        J_{ij}(\xt) = \frac{\partial D_i(\xt)}{\partial x_{t,j}} = \frac{\alpha_t^2 \tau^2}{\sigma_t^2 + N \alpha_t^2 \tau^2} = \frac{g}{N}.
    \end{align*}
    Setting $j = i$ gives $J_{ii}(\xt) = g/N$, and summing over the $N$ pixels $j \in R$ gives $\sum_{j \in R} J_{ij}(\xt) = N \cdot (g/N) = g$.
\end{proof}

\section{Benchmark and Baseline Details} \label{supp:bench}

\paragraph{DiT benchmark.} We train an unconditional DiT-B/2~\cite{peebles2023scalable} via flow matching~\cite{lipman2023flow} for $200$k steps with batch size $128$ on the $70{,}000$ FFHQ~\cite{karras2019style} face images at $256 \times 256$ resolution, encoded into $4 \times 32 \times 32$ latents using the Stable Diffusion autoencoder. Prior to training, an $80 \times 32$-pixel watermark is pasted into $7{,}000$ ($10\%$) of the training images, and $20$ clean training images are duplicated $100$ times each. After training, we generate $12{,}000$ samples using $50$ Euler steps. To assign ground-truth labels, we compute the maximum normalized cross-correlation (NCC) of each generated sample against the watermark template, which scores at least $0.979$ on watermarked training images and at most $0.28$ on clean training images. A sample is labeled \emph{watermarked (WM)} if its watermark NCC exceeds $0.6$, using the exact watermark footprint at the detected location as its ground-truth mask. A sample within a root-mean-square pixel distance of $0.05$ from any duplicated training image is labeled a \emph{near-copy} with an all-ones mask, whereas a sample with watermark NCC below $0.3$ and distance above $0.1$ from all duplicated images is labeled \emph{non-memorized (N)} with an empty mask. Among the $12{,}000$ generated samples, $548$ ($4.6\%$) are watermarked, $139$ are near-copies, and $10{,}527$ are non-memorized; the remaining $786$ samples ($6.6\%$) fall between these thresholds and are excluded to ensure unambiguous ground-truth labels. To match the scale of the Stable Diffusion benchmarks, we retain the first $300$ watermarked samples, the first $300$ non-memorized samples, and all $139$ near-copies. As with the Stable Diffusion prompts, samples whose index among the $12{,}000$ generations is divisible by $5$ are assigned to the development split ($58$ watermarked, $59$ non-memorized, and $37$ near-copy samples), and the remainder form the test split.

\paragraph{Computing the JaRS map for DiT.} Because the DiT is trained with flow matching rather than noise prediction, it parameterizes the velocity field $\x{v}_{\x{\theta}}(\xt, t)$. Using the equivalent noise prediction $\x{\epsilon}_{\x{\theta}}(\xt, t) = \xt + \alpha_t \x{v}_{\x{\theta}}(\xt, t)$, \cref{eq:xhat} yields the clean-data estimate $\xhat = \xt - \sigma_t \x{v}_{\x{\theta}}(\xt, t)$. The scaled denoiser is therefore $D(\xt) = \alpha_t (\xt - \sigma_t \x{v}_{\x{\theta}}(\xt, t))$, whose Jacobian (analogous to \cref{eq:jac}) is

\begin{align*}
    J(\xt) = \alpha_t \left(I - \sigma_t \frac{\partial \x{v}_{\x{\theta}}(\xt, t)}{\partial \xt}\right)
\end{align*}

At the initial sampling step ($t = 1$), $\alpha_t = 0$, so $J(\xt) = 0$ and the JaRS score satisfies $s_t(i) = C$ at every pixel (\cref{fig:supp:traj}). For Stable Diffusion, we note that the finite difference in \cref{eq:fd} differentiates only the text-conditional noise prediction $\x{\epsilon}_{\x{\theta}}(\cdot, t, y)$ and not the unconditional branch.

\paragraph{Evaluation protocol.} Following the implementation of \citet{kimlee2026}, every localization map is bilinearly upsampled from latent resolution to the $256 \times 256$ resolution of the ground-truth masks, and the Stage 1 summary statistic $z_d$ is computed on the upsampled map. After clipping at $0$, applying $\log(1 + x)$, and min-max normalizing across the evaluated split, the binarization threshold is selected from $1{,}001$ uniformly spaced levels in $[0, 1]$ to maximize mIoU. In the protocol of \citet{kimlee2026}, the min-max normalization bounds and the binarization threshold are determined on the evaluated test split itself. To verify that the results in \cref{tab:main} are not sensitive to test-split thresholding, we also evaluated fixing the min-max bounds and the binarization threshold on the balanced development split of each setting and applying them to the test split. Under this fixed-threshold evaluation, the test mIoU of JaRS+ changes by at most $0.001$ across all settings and by at most $0.013$ for the baselines, leaving every ranking in \cref{tab:main} unchanged. We therefore retain the standard protocol of \citet{kimlee2026} for direct comparability. Across the four models, the balanced test splits comprise $278$ (TV), $278 + 278$ (TV+N), and $3 \times 186$ (All) instances for SD v1.4; $471$ (TV) and $400 + 400$ (TV+N) for SD v2.1; $402$ (TV) and $3 \times 117$ (All) for Realistic Vision; and $242$ (WM), $241 + 241$ (WM+N), and $3 \times 102$ (All) for the DiT.

\paragraph{Baselines.} For Realistic Vision, where \citet{kimlee2026} do not specify a reference checkpoint, we use SD v1.1 as the checkpoint reference $\tilde{\x{\theta}}$. On the DiT, we saved intermediate checkpoints at $5$k, $20$k, $50$k, $100$k, and $150$k training steps; the $5$k-step checkpoint achieves the highest development mIoU for both $\Delta h_{\tilde\theta}$ and $\Delta s_{\tilde\theta}$ (in both their two-stage and published configurations; App.~\ref{supp:ref}) and is used throughout. In the two-stage comparison (`+') of \cref{tab:main}, every method selects its hyperparameters from the candidate sets (\cref{sec:method:jrsplus}): $\ell \in \{5, 10, \dots, 45, 49\}$; $d \in \{0, 5, \dots, 45, 49\}$ and $\phi \in \{\text{mean}, Q_{0.5}, Q_{0.9}, Q_{0.99}\}$; and $\tau$, selected among $37$ quantiles ($5\%$ to $95\%$) of $z_d$. Ties are broken in favor of the earlier step, and then in the order $Q_{0.9}$, mean, $Q_{0.5}$, $Q_{0.99}$. \cref{tab:supp:hparams} lists the selected $(\ell, d, \phi)$ for all methods. On the DiT, we also include step $48$ ($\sigma_t / \alpha_t = 0.042$) as an additional candidate for $\ell$ and $d$ to match the final-step noise-to-signal ratio of Stable Diffusion ($\sigma_t / \alpha_t = 0.041$, vs.\ $0.020$ at DiT step $49$), and likewise use step $48$ when evaluating the published final-step configuration of \citet{kimlee2026} (App.~\ref{supp:published}).

\begin{table*}[t]
    \centering
    \small
    \setlength{\tabcolsep}{4pt}
    \begin{tabular}{@{}llccccccccccccc@{}}
        \toprule
        & & \multicolumn{3}{c}{SD v1.4} & \multicolumn{3}{c}{SD v2.1} & \multicolumn{3}{c}{Realistic Vision} & \multicolumn{3}{c}{DiT (unconditional)} \\
        \cmidrule(lr){3-5} \cmidrule(lr){6-8} \cmidrule(lr){9-11} \cmidrule(l){12-14}
        Method & Reference & $\ell$ & $d$ & $\phi$ & $\ell$ & $d$ & $\phi$ & $\ell$ & $d$ & $\phi$ & $\ell$ & $d$ & $\phi$ \\
        \midrule
        Raw diagonal+ & -- & 40 & 10 & mean & 40 & 5 & $Q_{0.9}$ & 40 & 10 & mean & 40 & 30 & $Q_{0.99}$ \\
        $\Delta h_\varnothing$+ & null prompt & 49 & 49 & $Q_{0.9}$ & 49 & 30 & $Q_{0.5}$ & 49 & 20 & $Q_{0.5}$ & \multicolumn{3}{c}{\emph{not available}} \\
        $\Delta h_{\tilde\theta}$+ & checkpoint & 49 & 5 & $Q_{0.99}$ & 49 & 35 & mean & 49 & 5 & $Q_{0.9}$ & 40 & 30 & $Q_{0.99}$ \\
        $\Delta s_\varnothing$+ & null prompt & 45 & 49 & $Q_{0.9}$ & 45 & 35 & mean & 49 & 20 & $Q_{0.5}$ & \multicolumn{3}{c}{\emph{not available}} \\
        $\Delta s_{\tilde\theta}$+ & checkpoint & 45 & 0 & mean & 49 & 0 & $Q_{0.5}$ & 49 & 5 & $Q_{0.99}$ & 20 & 25 & $Q_{0.99}$ \\
        \midrule
        JaRS+ (ours) & -- & 15 & 5 & $Q_{0.99}$ & 15 & 5 & $Q_{0.9}$ & 35 & 5 & $Q_{0.99}$ & 35 & 25 & $Q_{0.99}$ \\
        \bottomrule
    \end{tabular}
    \caption{\textbf{Hyperparameters selected on the development split for all two-stage methods in \cref{tab:main}}: the Stage 2 localization step $\ell$, the Stage 1 detection step $d$, and the Stage 1 summary statistic $\phi$, selected using the candidate sets and rules in App.~\ref{supp:bench}. Step indices refer to the $50$-step sampler ($0$ to $49$). The single-stage JaRS row in \cref{tab:main} uses the same $\ell$ as JaRS+. On the DiT, checkpoint contrasts use the $5$k-step intermediate checkpoint as the reference $\tilde{\x{\theta}}$.}
    \label{tab:supp:hparams}
\end{table*}

\section{Alternative Row Statistics} \label{supp:l1}

\begin{figure*}[t]
    \centering
    \includegraphics[width=\textwidth]{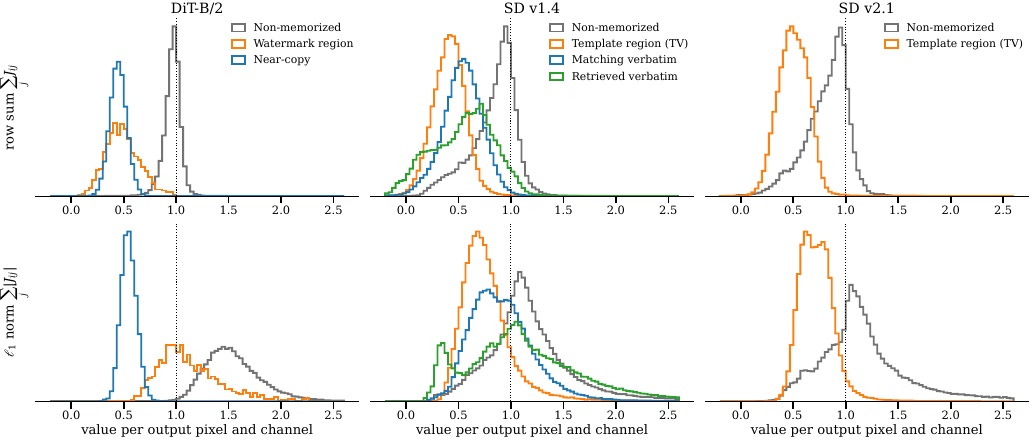}
    \caption{\textbf{Per-channel row sum $\sum_j J_{ij}$ (top) and $\ell_1$ row norm $\sum_j |J_{ij}|$ (bottom) of the full denoiser Jacobian} at the localization step of each model (DiT-B/2 at step $35$, SD v1.4 and SD v2.1 at step $15$), across all pixels of non-memorized images and whole-image copies, and inside the copied region of watermarked samples and template verbatims. The development images are those of \cref{sec:exp:rows}, together with $10$ template verbatims and $10$ non-memorized images of SD v2.1. The vertical dotted line at $1$ marks the row sum of a mean-preserving averaging filter.}
    \label{fig:supp:l1}
\end{figure*}

\begin{table}[t]
    \centering
    \small
    \setlength{\tabcolsep}{3.5pt}
    \begin{tabular}{@{}llccc@{}}
        \toprule
        Model & Memorized pixels & \begin{tabular}[c]{@{}c@{}}Per-channel\\row sum\\(JaRS)\end{tabular} & \begin{tabular}[c]{@{}c@{}}Full\\row sum\end{tabular} & \begin{tabular}[c]{@{}c@{}}$\ell_1$\\row norm\end{tabular} \\
        \midrule
        \multirow{2}{*}{DiT-B/2} & Watermark & \textbf{0.998} & 0.989 & 0.900 \\
        & Near-copy & \textbf{1.000} & \textbf{1.000} & \textbf{1.000} \\
        \midrule
        \multirow{3}{*}{SD v1.4} & Template (TV) & \textbf{0.980} & 0.975 & 0.908 \\
        & Matching verbatim & \textbf{0.917} & 0.868 & 0.772 \\
        & Retrieved verbatim & \textbf{0.862} & 0.827 & 0.574 \\
        \midrule
        SD v2.1 & Template (TV) & \textbf{0.955} & 0.941 & 0.934 \\
        \bottomrule
    \end{tabular}
    \caption{\textbf{Pixel-level AUROC between memorized and non-memorized pixels} for the JaRS map (\cref{eq:score}) and two variants, on the development images of \cref{fig:supp:l1}. Memorized pixels are those inside the copied region (all pixels for whole-image copies); non-memorized pixels are all pixels of the non-memorized development images. The full row sum also sums the Jacobian entries of the other latent channels, and the $\ell_1$ row norm replaces each per-channel row sum by the $\ell_1$ norm of the row.}
    \label{tab:supp:l1}
\end{table}

An alternative to the signed row sum is the $\ell_1$ norm $\sum_j |J_{(c,i),(c,j)}|$ of the row, which vanishes if and only if the entire row is zero. We evaluated this variant on the full Jacobians of \cref{sec:exp:rows}, together with $20$ development images of SD v2.1 ($10$ template verbatims and $10$ non-memorized images) at its localization step $\ell = 15$, replacing each row sum in \cref{eq:score} by the $\ell_1$ norm. On SD v2.1, the row sums behave as on the other two models: their median is $0.86$ on non-memorized images and $0.89$ outside the copied template, and falls to $0.50$ inside the template region. The $\ell_1$ norm, in contrast, does not concentrate around a common value on non-memorized images (\cref{fig:supp:l1}, bottom; median $1.51$ on the DiT and $1.13$ and $1.12$ on SD v1.4 and v2.1) and overlaps more with the memorized pixels. \cref{tab:supp:l1} confirms that the $\ell_1$ variant separates memorized from non-memorized pixels less accurately than the row sum for every model and category except the DiT near-copies, where both are perfect. The gap is largest on the whole-image copies of SD v1.4 and on the DiT watermark. We therefore use the signed row sum throughout.

In addition to being signed, the row sum in \cref{eq:score} is restricted to a single channel: for an output coordinate $(c, i)$, it sums $J_{(c, i), (c, j)}$ over the input pixels $j$ for the same channel $c$. A second alternative is the \emph{full row sum} $\sum_{c'=1}^{C} \sum_{j=1}^{HW} J_{(c,i),(c',j)}$, which aggregates across all input channels and requires only a single Jacobian-vector product with the all-ones vector $\one$ rather than $C$ channel-wise products. As shown in \cref{tab:supp:l1}, the full row sum matches the per-channel row sum on DiT near-copies, but separates memorized from non-memorized pixels less accurately across all other categories, particularly on SD v1.4 matching verbatims ($0.868$ vs.\ $0.917$). We therefore adopt the per-channel row sum.

One possible reason why the per-channel row sum is a reasonable choice is that it does not depend on how the latent channels are mixed. For a pixel $i$, collect the channel sums in a matrix $M_i \in \R^{C \times C}$ with $M_i[c, c'] = (J(\xt)\one_{c'})_{(c,i)}$. The JaRS score is then $s_t(i) = C - \operatorname{tr}(M_i)$, and the full row sum gives $C - \one^\top M_i \one$. If the latent is written in a different channel basis, mixing the channels at every pixel by the same invertible matrix $A$, then $M_i$ of the Jacobian becomes $A M_i A^{-1}$. This leaves the trace unchanged, but in general changes $\one^\top M_i \one$. Since the channels of the autoencoder latent have no natural preferred basis, this invariance seems a desirable property. Whether this invariance explains the results in \cref{tab:supp:l1} is left for future work.

\section{Ablation of Stage 1} \label{supp:detect}

\begin{table*}[t]
    \centering
    \small
    \setlength{\tabcolsep}{5pt}
    \begin{tabular}{@{}lcccccccccc@{}}
        \toprule
        & \multicolumn{3}{c}{SD v1.4} & \multicolumn{2}{c}{SD v2.1} & \multicolumn{2}{c}{Realistic Vision} & \multicolumn{3}{c}{DiT (unconditional)} \\
        \cmidrule(lr){2-4} \cmidrule(lr){5-6} \cmidrule(lr){7-8} \cmidrule(l){9-11}
        Variant & TV & TV+N & All & TV & TV+N & TV & All & WM & WM+N & All \\
        \midrule
        JaRS+ (tuned $\tau$, \cref{tab:main}) & 0.969 & 0.983 & 0.931 & 0.980 & 0.989 & 0.944 & 0.850 & 0.819 & 0.895 & 0.928 \\
        JaRS+ ($\tau$ at $5\%$ dev FPR) & 0.969 & 0.950 & 0.920 & 0.980 & 0.959 & 0.944 & 0.850 & 0.830 & 0.894 & 0.921 \\
        JaRS (no Stage 1) & 0.969 & 0.619 & 0.637 & 0.980 & 0.571 & 0.949 & 0.612 & 0.836 & 0.873 & 0.909 \\
        \midrule
        \emph{JaRS with oracle Stage 1} & 0.969 & 0.984 & 0.960 & 0.980 & 0.990 & 0.949 & 0.945 & 0.836 & 0.918 & 0.946 \\
        \bottomrule
    \end{tabular}
    \caption{\textbf{Ablation of Stage 1 on the test splits (mIoU).} All rows use the $(\ell, d, \phi)$ of JaRS+ in \cref{tab:supp:hparams}. `$\tau$ at $5\%$ dev FPR' sets $\tau$ to the $95$th percentile of $z_d$ on the non-memorized development images only; `JaRS' returns the map at step $\ell$ for every image; the oracle passes exactly the memorized images and is an upper bound for Stage 1.}
    \label{tab:supp:detect}
\end{table*}

\cref{tab:supp:detect} compares JaRS+ against three Stage 1 variants. First, the detection threshold $\tau$ can be calibrated without access to memorized examples by setting it to the $95$th percentile of $z_d$ on the non-memorized development images (fixing a $5\%$ development false-positive rate); this unsupervised threshold remains within $0.033$ mIoU of the tuned threshold across all settings. Second, omitting Stage 1 entirely alters mIoU by at most $0.017$ on TV and WM, where every image contains memorized content, but causes a sharp drop on settings that include non-memorized images---particularly for text-to-image models (e.g., from $0.983$ to $0.619$ on TV+N for SD v1.4). Third, pairing the JaRS maps with an oracle Stage 1 detector raises mIoU on the All setting of Realistic Vision from $0.850$ to $0.945$, nearly matching $\Delta h_\varnothing$ ($0.951$). This indicates that the performance gap on this setting stems from image-level detection in Stage 1 rather than the quality of the Stage 2 localization map.

\section{Baselines in Their Published Configuration} \label{supp:published}

In \cref{tab:main}, the baselines are evaluated in the two-stage form of JaRS+, with its hyperparameters selected on the development split. For completeness, \cref{tab:supp:published} reports the maps of \citet{kimlee2026} in their original published configuration: a single map evaluated at the final sampling step (step $48$ on the DiT to match the signal to noise ratio. Refer to App.~\ref{supp:bench}), without Stage 1 filtering. JaRS and JaRS+ are unchanged from \cref{tab:main}.

We also include the Bright Ending map of \citet{chen2025bright}. Following \citet[App.~F.3]{kimlee2026}, we compute this map as the cross-attention probability of the end-of-text token at the final sampling step, averaged across attention heads and the two $64 \times 64$ cross-attention layers in the first down-sampling block of the UNet. We evaluate it on the same sampling trajectories and test splits as all other methods, without Stage 1. Because Bright Ending relies on cross-attention to a text prompt, it is not applicable to the unconditional DiT.


\begin{table*}[t]
    \centering
    \small
    \setlength{\tabcolsep}{5pt}
    \begin{tabular}{@{}llcccccccccc@{}}
        \toprule
        & & \multicolumn{3}{c}{SD v1.4} & \multicolumn{2}{c}{SD v2.1} & \multicolumn{2}{c}{Realistic Vision} & \multicolumn{3}{c}{DiT (unconditional)} \\
        \cmidrule(lr){3-5} \cmidrule(lr){6-7} \cmidrule(lr){8-9} \cmidrule(l){10-12}
        Method & Reference & TV & TV+N & All & TV & TV+N & TV & All & WM & WM+N & All \\
        \midrule
        Bright Ending~\cite{chen2025bright} & -- & 0.755 & 0.500 & 0.560 & 0.933 & 0.954 & 0.779 & 0.591 & \multicolumn{3}{c}{\emph{not available}} \\
        Raw diagonal~\cite{kimlee2026} & -- & 0.594 & 0.500 & 0.524 & 0.652 & 0.500 & 0.660 & 0.530 & 0.261 & 0.500 & 0.640 \\
        $\Delta h_\varnothing$~\cite{kimlee2026} & null prompt & 0.900 & 0.950 & \textbf{0.945} & 0.945 & 0.887 & 0.930 & \textbf{0.951} & \multicolumn{3}{c}{\emph{not available}} \\
        $\Delta h_{\tilde\theta}$~\cite{kimlee2026} & checkpoint & 0.913 & 0.903 & 0.884 & 0.948 & 0.876 & 0.936 & 0.880 & 0.439 & 0.500 & 0.666 \\
        $\Delta s_\varnothing$~\cite{kimlee2026} & null prompt & 0.826 & 0.901 & 0.898 & 0.787 & 0.809 & 0.876 & 0.903 & \multicolumn{3}{c}{\emph{not available}} \\
        $\Delta s_{\tilde\theta}$~\cite{kimlee2026} & checkpoint & 0.864 & 0.730 & 0.653 & 0.921 & 0.877 & 0.860 & 0.626 & 0.376 & 0.501 & 0.523 \\
        \midrule
        JaRS (ours) & -- & 0.969 & 0.619 & 0.637 & 0.980 & 0.571 & 0.949 & 0.612 & 0.836 & 0.873 & 0.909 \\
        JaRS+ (ours) & -- & \textbf{0.969} & \textbf{0.983} & 0.931 & \textbf{0.980} & \textbf{0.989} & \textbf{0.944} & 0.850 & \textbf{0.819} & \textbf{0.895} & \textbf{0.928} \\
        \bottomrule
    \end{tabular}
    \caption{\textbf{Pixel-level localization on the test splits (mIoU) with baselines in their published configuration}: a single map evaluated at the final sampling step (step $48$ on the DiT), without Stage 1, using $16$ Hutchinson probes and the checkpoint references of \cref{tab:main}. JaRS and JaRS+ are repeated from \cref{tab:main} for comparison. Bold indicates the best result in each column (excluding JaRS).}
    \label{tab:supp:published}
\end{table*}

Compared against the baselines in their published configurations, JaRS+ achieves the highest mIoU in eight of the ten settings, trailing $\Delta h_\varnothing$ only on the All splits of SD v1.4 and Realistic Vision (as in \cref{tab:main}). On the purely memorized TV splits, the single-stage published configuration of $\Delta h_{\tilde\theta}$ outperforms its two-stage counterpart ($0.913$, $0.948$, and $0.936$ vs.\ $0.879$, $0.918$, and $0.895$ on SD v1.4, SD v2.1, and Realistic Vision), since Stage 1 filtering can only discard true-positive masks when every evaluation image is memorized; nevertheless, both configurations remain well below JaRS+. Bright Ending likewise falls short of JaRS+ across all evaluated settings.

Finally, we note that the mixed settings (TV+N, WM+N, and All) are particularly sensitive to how empty ground-truth masks are scored. Under the protocol of \citet{kimlee2026}, every method is binarized using a single global pixel threshold across all images in a split, and a non-memorized image (whose ground-truth mask is empty) receives an IoU of $1$ if its predicted mask is completely empty and $0$ if even one pixel exceeds the threshold. Consequently, a non-memorized image scores $1$ if and only if every pixel value lies below the global threshold.

Two-stage methods enforce this directly by zeroing out the map whenever Stage 1 rejects an image, whereas single-stage contrast maps such as $\Delta h_\varnothing$ rely on reference subtraction to suppress pixel values on non-memorized images. Conversely, when a single-stage map exhibits large pixel values on non-memorized images (such as Raw diagonal, Bright Ending on SD v1.4, or JaRS on smooth backgrounds), any threshold low enough to segment the copied regions also triggers false-positive pixels on non-memorized images, zeroing out their IoU. In extreme cases, the optimal threshold lies above every map value---predicting an empty mask for all images and yielding an exact mIoU of $0.500$ on the balanced TV+N and WM+N splits (e.g., Bright Ending on SD v1.4 TV+N and Raw diagonal across all models). As a result, performance differences across methods are amplified on the mixed splits and largely reflect how reliably each method produces an empty mask on non-memorized images.

\section{The JaRS Map Across Noise Levels} \label{supp:traj}

\begin{figure}[t]
    \centering
    \includegraphics[width=\columnwidth]{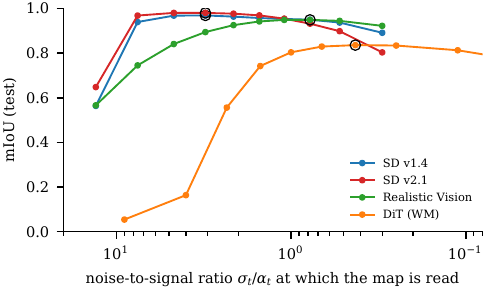}
    \caption{\textbf{Effect of noise level on localization accuracy.} Test mIoU of the JaRS map $s_t$ on locally memorized instances (TV for text-to-image models, WM for the DiT) plotted against the noise-to-signal ratio $\sigma_t/\alpha_t$ at which the map is evaluated. Black circles mark the localization step $\ell$ selected on the development split.}
    \label{fig:noise}
\end{figure}

\paragraph{Localization accuracy across noise levels.} \cref{fig:noise} plots the test mIoU of the JaRS map $s_t$ as a function of the noise-to-signal ratio $\sigma_t / \alpha_t$, which places the variance-preserving schedule of Stable Diffusion and the flow-matching schedule of the DiT on a unified axis. The empirical curves follow the trend predicted by \cref{prop:smooth}. At the highest noise levels ($\sigma_t / \alpha_t \gg 1$), the denoiser discards its noisy input everywhere ($g \to 0$), so on SD v1.4 and v2.1 the map scores $0.563$ and $0.648$ (the trivial accuracy of an all-ones mask). As noise decreases, mIoU rises sharply to a broad, stable plateau centered around $\sigma_t / \alpha_t \approx 3$ for SD v1.4 and v2.1, and around $\sigma_t / \alpha_t \in [0.3, 1]$ for Realistic Vision and the DiT---before declining as $\sigma_t / \alpha_t \to 0$, where $g \to 1$ everywhere. Notably, at the final sampling step ($\sigma_t / \alpha_t \approx 0.04$), where \citet{kimlee2026} evaluate their contrast maps, a single model's Jacobian approaches the identity everywhere and the JaRS map returns to the trivial baseline ($0.563$ and $0.648$). This suggests that contrast-free localization works best at intermediate $\sigma_t / \alpha_t$ rather than at the final step: across all four models, the localization step $\ell$ chosen on the development split (circled in \cref{fig:noise}) lies on the plateau, and on Stable Diffusion the localization map is available after $15$ of the $50$ sampling steps.

\begin{figure*}[t]
    \centering
    \includegraphics[width=\textwidth]{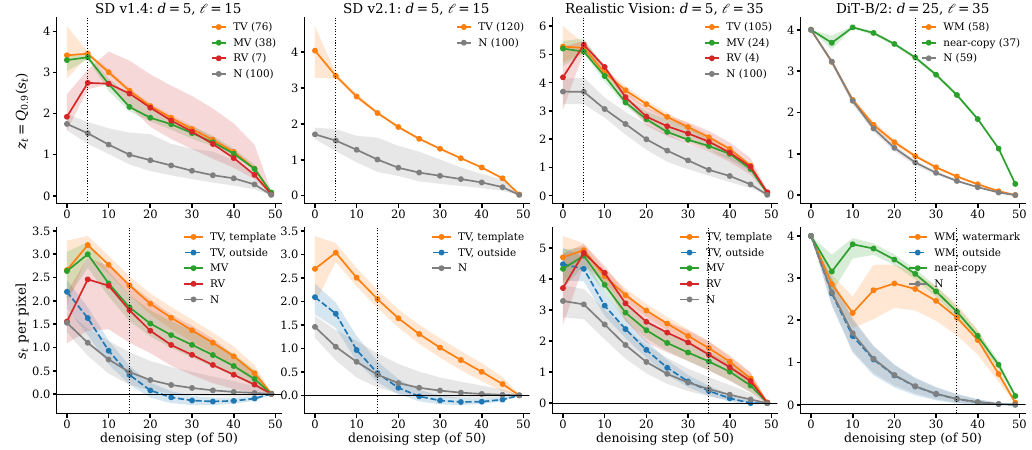}
    \caption{\textbf{Evolution of the JaRS map $s_t$ across the $50$-step sampling trajectory} on the development split. Curves and shaded bands show the median and interquartile range, with category counts in the legend. \emph{Top:} Image-level score $z_t = Q_{0.9}(s_t)$ per category. We plot $Q_{0.9}$ across all models for uniform comparison but note that Stage 1 of JaRS+ uses the tuned statistic $\phi$ in \cref{tab:supp:hparams}. \emph{Bottom:} Pixel-level scores $s_t$ inside vs.\ outside the copied region for locally memorized images (TV and WM), and across the full image for the remaining categories, computed over $1{,}024$ randomly sampled pixels per image. Vertical dotted lines indicate the Stage 1 detection step $d$ (top) and Stage 2 localization step $\ell$ (bottom) of JaRS+ (\cref{tab:main}).}
    \label{fig:supp:traj}
\end{figure*}

\paragraph{The JaRS map along the sampling trajectory.} \cref{fig:supp:traj} tracks the JaRS map $s_t$ across all $50$ denoising steps on development images, illustrating the two-timescale separation between early image-level detection and intermediate pixel-level localization that motivates JaRS+ (\cref{sec:method:jrsplus}).

\paragraph{Early image-level separation (Stage 1).} On all three text-to-image models (\cref{fig:supp:traj}, top), the image-level score $z_t = Q_{0.9}(s_t)$ is systematically higher on memorized trajectories than on non-memorized trajectories at every step prior to the end of sampling, with the widest separation occurring in the earliest steps. On SD v1.4, for example, the median $z_t$ on template verbatims is $3.46$ at step $5$ versus $1.52$ on non-memorized images (compared to $1.07$ vs.\ $0.42$ by step $40$). Retrieved verbatims begin at a lower median at step $0$ ($1.92$ on SD v1.4 and $4.19$ on Realistic Vision, compared to $3.42$ and $5.28$ for template verbatims) before catching up to the other memorized categories by step $5$ on Realistic Vision and step $10$ on SD v1.4. This trajectory behavior explains why $z_t$ separates memorized from non-memorized images less well at step $0$ than at step $5$ on all three text-to-image models. At the final step ($49$), as the noise level $\sigma_t / \alpha_t$ approaches zero, the median $z_t$ of every category collapses to at most $0.12$, consistent with \cref{prop:smooth}.

\paragraph{Intermediate pixel-level separation (Stage 2).} The pixel-level curves (\cref{fig:supp:traj}, bottom) explain why accurate spatial localization requires waiting until an intermediate step $\ell > d$. In the earliest denoising steps, the unmemorized pixels \emph{outside} the copied template of a TV image exhibit nearly the same elevated JaRS scores as the pixels \emph{inside} the template: on SD v1.4 at step $0$, the median score is $2.19$ outside the template versus $2.66$ inside. Thus, at very high noise, a memorized prompt elevates the JaRS map across the entire image, which provides a strong signal for image-level detection but poor contrast for segmenting the copied region. By step $\ell = 15$ on SD v1.4 and SD v2.1, the pixels outside the template relax to the baseline level of ordinary non-memorized images (medians of $0.41$ and $0.48$, matching $0.47$ and $0.44$ on non-memorized images), whereas the pixels inside the copied template remain elevated ($2.33$ and $2.05$), creating a clear spatial boundary. On Realistic Vision, the unmemorized background relaxes more gradually and reaches the non-memorized level only around step $30$ ($0.70$ vs.\ $0.67$), consistent with its later selected localization step $\ell = 35$. Beyond step $20$, the median score outside the template dips slightly below zero on SD v1.4 and SD v2.1 (reaching $-0.16$), where sharpening side lobes cause the per-channel row sums to slightly exceed $1$; these negative values are clipped at $0$ in the evaluation protocol (\cref{sec:exp:setup}).

\paragraph{Trajectory dynamics on the unconditional DiT.} On the unconditional DiT (\cref{fig:supp:traj}, rightmost column), global and local memorization emerge on different timescales. Whole-image near-copies separate from non-memorized samples almost immediately: the AUROC of $z_t$ between near-copies and non-memorized samples reaches $0.960$ at step $5$ and $1.000$ from step $15$ onward. In contrast, the planted watermark covers only $3.8\%$ of the image area on average, so it barely shifts the $90$th spatial percentile $z_t$ (at step $30$, the median $z_{30}$ is $0.68$ on watermarked samples versus $0.54$ on non-memorized samples). Within each watermarked sample, pixel-level separation between the watermark and the unmemorized face background develops only at intermediate noise: the mean within-image pixel AUROC rises from $0.547$ at step $0$ and $0.769$ at step $10$ to $0.949$ at step $15$, and exceeds $0.986$ from step $20$ onward. Because no text prompt locks in the watermark at $t \approx 1$, image-level separation between watermarked and non-memorized samples peaks later, at step $30$ (AUROC $0.925$); JaRS+ selects $d = 25$ together with the $99$th percentile $Q_{0.99}$ on development data (\cref{sec:method:jrsplus}).

\section{Sensitivity of Checkpoint Contrasts to the Reference Checkpoint}
\label{supp:ref}

\begin{table}[t]
    \centering
    \small
    \resizebox{\columnwidth}{!}{%
    \begin{tabular}{@{}llcccc@{}}
        \toprule
        & & \multicolumn{4}{c}{Localization (mIoU)} \\
        \cmidrule(l){3-6}
        & Reference $\tilde{\x{\theta}}$ & Dev All & WM & WM+N & All \\
        \midrule
        \multirow{5}{*}{$\Delta h_{\tilde\theta}$} & $5$k ($2.5\%$) & \textbf{0.666} & 0.439 & 0.500 & 0.666 \\
        & $20$k ($10\%$) & 0.663 & 0.503 & 0.500 & 0.663 \\
        & $50$k ($25\%$) & 0.661 & 0.457 & 0.500 & 0.655 \\
        & $100$k ($50\%$) & 0.634 & 0.369 & 0.500 & 0.616 \\
        & $150$k ($75\%$) & 0.531 & 0.243 & 0.500 & 0.526 \\
        \midrule
        \multirow{5}{*}{$\Delta s_{\tilde\theta}$} & $5$k ($2.5\%$) & \textbf{0.507} & 0.376 & 0.501 & 0.523 \\
        & $20$k ($10\%$) & 0.473 & 0.461 & 0.504 & 0.482 \\
        & $50$k ($25\%$) & 0.423 & 0.477 & 0.506 & 0.434 \\
        & $100$k ($50\%$) & 0.402 & 0.511 & 0.515 & 0.409 \\
        & $150$k ($75\%$) & 0.395 & 0.539 & 0.530 & 0.402 \\
        \midrule
        \multicolumn{2}{@{}l}{\textbf{JaRS+ (no reference)}} & -- & \textbf{0.819} & \textbf{0.895} & \textbf{0.928} \\
        \bottomrule
    \end{tabular}}
    \caption{\textbf{Sensitivity of the checkpoint contrasts $\Delta h_{\tilde\theta}$ and $\Delta s_{\tilde\theta}$~\cite{kimlee2026} on the unconditional DiT} (trained for $200$k steps) across all saved intermediate checkpoints $\tilde{\x{\theta}}$ (fraction of training completed shown in parentheses), in their published configuration: at step $48$ ($\sigma_t/\alpha_t = 0.042$, matching the final step of Stable Diffusion) with $16$ Hutchinson probes and without Stage 1. The `Dev All' column is used to select the reference checkpoint (bold); the remaining columns report test-split performance. JaRS+ is as in \cref{tab:main}.}
    \label{tab:supp:ref}
\end{table}

\citet{kimlee2026} note that the reference checkpoint $\tilde{\x{\theta}}$ in $\Delta h_{\tilde\theta}$ must be ``relatively less overfitted'' than the target model $\x{\theta}$, cautioning that overly early checkpoints fail to capture general data-driven curvature while overly late checkpoints already exhibit memorization-driven curvature. For the unconditional DiT (trained for $200$k steps), we evaluated both $\Delta h_{\tilde\theta}$ and $\Delta s_{\tilde\theta}$ in their published configuration (step $48$; App.~\ref{supp:bench}) across all five intermediate checkpoints saved during training ($5$k, $20$k, $50$k, $100$k, and $150$k steps). \Cref{tab:supp:ref} reports the resulting development and test localization accuracy.

For localization, the earliest checkpoint ($5$k steps) achieves the highest development All mIoU for both contrasts and is therefore used in \cref{tab:main}. On WM alone, $\Delta h_{\tilde\theta}$ peaks at the $20$k checkpoint ($0.503$ mIoU, which is not identified by the development split) and degrades sharply as the reference checkpoint advances through training, dropping to $0.243$ at $150$k steps. Across all five checkpoints, neither $\Delta h_{\tilde\theta}$ nor $\Delta s_{\tilde\theta}$ separates watermarked samples from non-memorized samples on WM+N (reaching at most $0.530$ mIoU), and every checkpoint falls well short of JaRS+ across all three localization settings (by at least $0.262$ mIoU).

\section{Prompt-Level Memorization Detection}
\label{supp:promptdet}

On Stable Diffusion, the positive set consists of the memorized prompts cataloged by \citet{webster2023reproducible}: all $500$ prompts for SD v1.4 and the $219$ prompts labeled as verbatim copies for SD v2.1, while the negative set consists of the $500$ non-memorized prompts from the localization benchmark. Following \citet[Sec.~5.2]{kimlee2026}, each prompt is generated with four random seeds, and its score is the mean over the four seeds of the method's Stage 1 statistic $\phi$ of its map at step $d$, with $(d, \phi)$ taken from the localization setting of \cref{tab:main} and not re-tuned. The maps of \citet{kimlee2026} are computed with their official code. All metrics are reported on the held-out test split. Since $(d, \phi)$ are selected on the development split of the localization benchmark, we exclude the memorized test prompts that also appear there ($43$ on SD v1.4 and $34$ on SD v2.1), which leaves $357$ memorized vs.\ $400$ non-memorized prompts on SD v1.4, and $140$ vs.\ $400$ on SD v2.1. On the DiT, we score individual test samples ($344$ memorized watermark or near-copy samples vs.\ $241$ non-memorized samples), and the checkpoint contrasts use the $5$k-step reference.

\section{Computation: Differentiation Mode, Precision, and Cost}
\label{supp:cost}

In this section, we detail the practical computation of the JaRS map. A natural question is whether the row sums can be computed via reverse-mode automatic differentiation (backpropagation). Differentiating the scalar $\one_c^\top D(\xt)$ in reverse mode, however, yields the \emph{column sums} $J(\xt)^\top \one_c$ rather than the row sums $J(\xt)\one_c$. While the Bayes-optimal denoiser Jacobian (\cref{eq:cov}) is symmetric, the Jacobian of a trained neural network is generally asymmetric, making forward directional derivatives necessary to obtain the true row sums. In principle, these directional derivatives can be computed via either exact forward-mode automatic differentiation or central finite differences (\cref{eq:fd}); we adopt finite differences because batching all $C$ directions in forward-mode automatic differentiation exceeded GPU memory in our implementation. Crucially, the forward passes in \cref{eq:fd} must be evaluated in full single precision (\texttt{fp32}, with TensorFloat-32 disabled). We observed that reduced-precision convolutions introduce roughly a $20\%$ relative error in the finite-difference quotient on Stable Diffusion, whereas in \texttt{fp32} the central difference in \cref{eq:fd} matches exact forward-mode automatic differentiation to within a relative error of $10^{-4}$.

\begin{table*}[t]
    \centering
    \small
    \setlength{\tabcolsep}{5pt}
    \begin{tabular}{@{}lllccccccc@{}}
        \toprule
        & & & \multicolumn{4}{c}{Stable Diffusion} & & \multicolumn{2}{c}{DiT} \\
        \cmidrule(lr){4-7} \cmidrule(l){9-10}
        Method & Step & Precision & TFLOPs & SD v1.4 (s) & SD v2.1 (s) & RV (s) & & TFLOPs & Time (s) \\
        \midrule
        Raw diagonal & last & \texttt{fp16} / \texttt{fp32} & 16.6 & 0.49 & 0.43 & 0.50 & & 1.47 & 0.106 \\
        $\Delta h_\varnothing$ & last & \texttt{fp16} & 33.3 & 0.81 & 0.69 & 0.81 & & -- & -- \\
        $\Delta h_{\tilde\theta}$ & last & \texttt{fp16} / \texttt{fp32} & 33.3 & 0.99 & 0.86 & 0.99 & & 2.95 & 0.213 \\
        $\Delta s_\varnothing$ & last & \texttt{fp16} & 1.6 & 0.04 & 0.03 & 0.04 & & -- & -- \\
        $\Delta s_{\tilde\theta}$ & last & \texttt{fp16} / \texttt{fp32} & 1.6 & 0.05 & 0.04 & 0.05 & & 0.09 & 0.012 \\
        \midrule
        JaRS (one map) & $\ell$ & \texttt{fp32} & 6.4 & 0.46 & 0.41 & 0.46 & & 0.37 & 0.029 \\
        JaRS+ (both stages) & $d$, $\ell$ & \texttt{fp32} & 12.9 & 0.91 & 0.83 & 0.92 & & 0.74 & 0.058 \\
        JaRS+ (mean, broadest setting) & $d$, $\ell$ & \texttt{fp32} & 9.7--10.5 & 0.74 & 0.62 & 0.75 & & 0.62 & 0.049 \\
        \midrule
        \emph{Generating the image} & all & \texttt{fp16} / \texttt{bf16} & 80.3 & 1.80 & 1.69 & 1.84 & & 2.30 & 0.329 \\
        \bottomrule
    \end{tabular}
    \caption{\textbf{Computational cost per test image for each localization map} on a single NVIDIA A6000 GPU. FLOPs are measured with PyTorch's FLOP counter and are identical across the three Stable Diffusion models. Wall-clock times are medians over $20$ test instances with three repetitions each. Each baseline of \citet{kimlee2026} is timed for a single map at its published step (`last', which is step $49$ on Stable Diffusion and step $48$ on the DiT), whereas the two-stage variants in \cref{tab:main} evaluate up to two steps, as in JaRS+. On Stable Diffusion, baselines run in \texttt{fp16} using the official code of \citet{kimlee2026} with $16$ Hutchinson vector-Jacobian products; on the DiT, they run in \texttt{fp32} using $16$ Rademacher central-difference probes. JaRS always runs in \texttt{fp32}. Because JaRS+ computes the Stage 2 map only when Stage 1 detects memorization, we report both the cost for a detected image (`both stages') and the average cost across the broadest balanced test split (`mean', where Stage 1 accepts $63\%$, $50\%$, $62\%$, and $67\%$ of images on SD v1.4, SD v2.1, Realistic Vision, and the DiT). Full $50$-step image generation is included for reference.}
    \label{tab:supp:cost}
\end{table*}

\cref{tab:supp:cost} compares the computational cost of all methods in \cref{tab:main}. Computing a single JaRS map requires $2C = 8$ forward passes and no backward pass, whereas the raw diagonal requires the equivalent of roughly $21$ forward passes and $\Delta h_\varnothing$ and $\Delta h_{\tilde\theta}$ require roughly $41$ on Stable Diffusion, since each input-gradient backward pass costs about $1.2$ forward passes. Averaged over the broadest test split, JaRS+ uses $3.2$--$3.4\times$ fewer FLOPs than $\Delta h_\varnothing$ and $\Delta h_{\tilde\theta}$ on Stable Diffusion, and $4.8\times$ fewer than $\Delta h_{\tilde\theta}$ on the DiT. In wall-clock time on Stable Diffusion, the gap is narrower because the finite difference in \cref{eq:fd} runs in \texttt{fp32} while the official implementation of \citet{kimlee2026} runs in \texttt{fp16}: JaRS+ takes $0.62$--$0.75$\,s per image compared to $0.69$--$0.81$\,s for $\Delta h_\varnothing$ and $0.86$--$0.99$\,s for $\Delta h_{\tilde\theta}$, and is slightly slower than the raw diagonal at $0.43$--$0.50$\,s. When evaluated in \texttt{fp32} on Stable Diffusion, $\Delta h_\varnothing$ and $\Delta h_{\tilde\theta}$ require $2.8$--$3.8$\,s per image. On the DiT, where all methods are evaluated in \texttt{fp32}, JaRS+ takes $0.049$\,s versus $0.213$\,s for $\Delta h_{\tilde\theta}$. While the score contrasts $\Delta s_\varnothing$ and $\Delta s_{\tilde\theta}$ are the fastest methods at just two forward passes, they localize memorized regions substantially less accurately (\cref{tab:main}). Overall, every localization map requires only a fraction of the time needed to generate the image itself.

Selecting the hyperparameters $(\ell, d, \phi, \tau)$ on the development split is a one-time calibration cost per model. It requires evaluating JaRS maps across the candidate steps ($11$ steps on Stable Diffusion and $12$ on the DiT; App.~\ref{supp:bench}) for the $166$, $220$, $161$, and $132$ development images of SD v1.4, SD v2.1, Realistic Vision, and the DiT respectively. At the per-map runtimes in \cref{tab:supp:cost}, this calibration takes roughly $14$, $16$, $13$, and $1$ minutes on a single GPU, in addition to $5$, $6$, $5$, and $1$ minutes to generate the development trajectories. By comparison, the published configuration of \citet{kimlee2026} requires no development split on Stable Diffusion, whereas on the DiT the checkpoint contrasts require saving intermediate training checkpoints and evaluating five candidate references on the development split (\cref{sec:exp:setup}).

\section{Region-Restricted Memorization Mitigation}
\label{supp:mitig}

In this section, we use the JaRS map to suppress memorization strictly within the localized region while preserving the rest of the image unchanged. We first describe our region-restricted sampling procedure (\cref{sec:method:mitigation}) and then compare it against whole-image mitigation on Stable Diffusion v1.4 and v2.1 (\cref{sec:exp:mitigation}).

\subsection{Method} \label{sec:method:mitigation}

Existing inference-time mitigations modify the prompt conditioning or initial noise and regenerate the entire image. For instance, random token addition (RTA)~\cite{somepalli2023understanding} inserts random tokens into the prompt, \citet{wen2024detecting} optimize the prompt embedding to minimize the magnitude of the text-conditional noise prediction at the first sampling step, and other methods perturb the initial noise~\cite{jeon2025sail}. While these approaches effectively disrupt the memorized template, regenerating the entire image inevitably alters the unmemorized regions as well.

Even when a localization map is used, as in Bright Ending~\cite{chen2025bright}---which weights the first-step optimization objective of \citet{wen2024detecting} using the localization map, the prompt embedding is still optimized at the original initial noise and used to regenerate the whole image. Weighting the optimization loss changes which pixels drive the prompt update, but because the modified prompt conditions the entire image during sampling, nothing prevents the unmemorized background from changing.

In contrast, our procedure applies the spatial mask directly in the sampler rather than in the loss, preserving the unmemorized region by construction and remaining agnostic to the choice of prompt mitigation. Let $R \in \{0, 1\}^{h \times w}$ be a binary mask on the latent grid, obtained by thresholding the JaRS map $s_{t_\ell}$ and dilating the result by $2$ latent pixels across all channels. Let $\x{x}_1$ denote the initial noise of the original generation, $\xo$ its final clean latent, and $\tilde y$ the mitigated prompt conditioning. To break the seed-specific memorization trigger inside $R$ while keeping the surrounding trajectory intact, we start a new sampling run from the composite latent
\begin{align} \label{eq:mitig}
    \x{x}_1^R = R \odot \x{\xi} + (1 - R) \odot \x{x}_1, \qquad \x{\xi} \sim \mathcal{N}(0, I),
\end{align}
where fresh Gaussian noise $\x{\xi}$ is drawn inside $R$ and the original noise $\x{x}_1$ is kept outside $R$. At each of the $50$ DDIM steps, we evaluate the classifier-free guided noise prediction $\x{\epsilon}_\varnothing + w(\x{\epsilon}_{\tilde y} - \x{\epsilon}_\varnothing)$ under the mitigated conditioning $\tilde y$ with guidance scale $w$. After each step to noise level $t'$, the latent outside $R$ is replaced by the forward-noised original latent $\alpha_{t'} \xo + \sigma_{t'} \x{\epsilon}$ using one fixed noise realization $\x{\epsilon}$, following the known-region replacement strategy of RePaint~\cite{lugmayr2022repaint} without resampling steps. Consequently, the noise prediction outside $R$ has no effect on the final result, and after the last step the latent outside $R$ is identical to $\xo$. Inside $R$, the network sees the preserved context outside $R$ at every step, generating new content that blends coherently with the rest of the image.

When $\tilde y$ is obtained via prompt-embedding optimization~\cite{wen2024detecting, chen2025bright}, one must choose the initial latent at which the embedding is optimized. Because our sampler starts from $\x{x}_1^R$, we optimize $\tilde y$ at $\x{x}_1^R$ by default, ensuring that every method calibrates its prompt embedding at the exact latent from which it subsequently samples. Because the original initial noise $\x{x}_1$ in the benchmark is a seed that triggered memorization, replacing the noise inside $R$ to form $\x{x}_1^R$ already partially lowers the initial text-conditional spike at the first step; in \cref{sec:exp:mitigation}, we therefore also compare against optimizing $\tilde y$ at $\x{x}_1$, which induces a larger update to the prompt embedding. Aside from computing the JaRS map at step $\ell$ ($8$ forward passes; App.~\ref{supp:cost}), our procedure incurs the exact same computational cost as whole-image mitigation.

\subsection{Experimental Setup and Results} \label{sec:exp:mitigation}

\begin{table*}[t]
    \centering
    \small
    \setlength{\tabcolsep}{4pt}
    \begin{tabular}{@{}lllccccc@{}}
        \toprule
        Mitigation $\tilde y$ & Mask in loss & Sampling & \begin{tabular}[c]{@{}c@{}}Copied\\(MSE) $\downarrow$\end{tabular} & \begin{tabular}[c]{@{}c@{}}Copied\\(SSCD) $\downarrow$\end{tabular} & LPIPS$_{\bar R}$ $\downarrow$ & LPIPS$_{\bar T}$ $\downarrow$ & CLIP $\uparrow$ \\
        \midrule
        None & -- & Original image & 0.975 & 0.996 & 0.000 & 0.000 & 29.0 \\
        Original prompt & -- & Region $R$ & 0.694 & 0.705 & 0.054 & 0.084 & 29.2 \\
        Null prompt & -- & Region $R$ & 0.000 & 0.004 & 0.178 & 0.265 & 13.6 \\
        \midrule
        \multirow{3}{*}{RTA~\cite{somepalli2023understanding}} & -- & Whole image & 0.432 & 0.590 & 0.444 & 0.437 & 25.0 \\
        & -- & Whole image, fresh noise & 0.471 & 0.507 & 0.538 & 0.529 & 25.1 \\
        & -- & Region $R$ (ours) & 0.432 & 0.475 & 0.094 & 0.148 & 26.8 \\
        \midrule
        \multirow{5}{*}{Wen et al.~\cite{wen2024detecting}} & -- & Whole image & 0.205 & 0.234 & 0.708 & 0.689 & 20.4 \\
        & -- & Whole image, fresh noise & 0.248 & 0.270 & 0.787 & 0.761 & 19.3 \\
        & -- & Region $R$ (ours) & 0.122 & 0.147 & 0.147 & 0.224 & 24.0 \\
        & -- & \emph{Region $R$, no fresh noise} & 0.345 & 0.392 & 0.120 & 0.180 & 23.9 \\
        & -- & \emph{Region $R$, embedding at $\x{x}_1$} & 0.090 & 0.122 & 0.158 & 0.239 & 22.2 \\
        \midrule
        \multirow{6}{*}{Bright Ending~\cite{chen2025bright}} & none & Whole image & 0.137 & 0.158 & 0.707 & 0.694 & 23.0 \\
        & JaRS mask $R$ & Whole image & 0.101 & 0.129 & 0.700 & 0.691 & 21.9 \\
        & Bright Ending map & Whole image$^\ast$ & 0.108 & 0.129 & 0.687 & 0.680 & 22.8 \\
        & none & Region $R$ & 0.216 & 0.230 & 0.115 & 0.188 & 26.7 \\
        & JaRS mask $R$ & Region $R$ & 0.198 & 0.212 & 0.124 & 0.202 & 25.3 \\
        & Bright Ending map & Region $R$ & 0.191 & 0.201 & 0.125 & 0.204 & 26.1 \\
        \midrule
        RTA~\cite{somepalli2023understanding} & -- & Region $R$ from $\Delta h_{\tilde\theta}$ & 0.471 & 0.525 & 0.082 & 0.146 & 26.9 \\
        Wen et al.~\cite{wen2024detecting} & -- & Region $R$ from $\Delta h_{\tilde\theta}$ & 0.122 & 0.173 & 0.149 & 0.252 & 23.9 \\
        \bottomrule
    \end{tabular}
    \caption{\textbf{Whole-image vs.\ region-restricted mitigation on the $278$ test TV instances of Stable Diffusion v1.4} at the default mitigation strength. `Copied' reports the fraction of generated images that still reproduce the ground-truth template under the benchmark's MSE criterion or under SSCD similarity. LPIPS$_{\bar R}$ and LPIPS$_{\bar T}$ measure perceptual change relative to the original unmitigated generation outside the latent mask $R$ and outside the ground-truth template $T$, respectively, while CLIP measures image--prompt alignment. `Region $R$' applies our region-restricted sampling procedure (\cref{sec:method:mitigation}) using the JaRS mask unless $\Delta h_{\tilde\theta}$ is specified. `Whole image, fresh noise' runs whole-image mitigation starting from the fresh noise realization $\x{\xi}$ used inside $R$. `Mask in loss' indicates the spatial weighting applied to the optimization objective of Bright Ending~\cite{chen2025bright}, where $^\ast$ denotes its published configuration. Rows in italics are ablations of our procedure.}
    \label{tab:mitig}
\end{table*}

\paragraph{Benchmarks and masks.} We evaluate mitigation on the $278$ test TV instances of SD v1.4 and the $471$ test TV instances of SD v2.1. Following the localization benchmark, all images are generated with $50$ DDIM steps~\cite{song2021denoising} and classifier-free guidance scale $w = 7.5$. To construct the binary mask $R$, we threshold the JaRS map at step $\ell = 15$ at latent resolution and dilate the resulting mask by $2$ latent pixels; the binarization threshold is selected among $91$ quantiles ($5\%$ to $95\%$) of the development maps to maximize latent-resolution mIoU against the ground-truth template masks on the development TV split. On SD v1.4, the resulting dilated JaRS mask covers $99.9\%$ of the ground-truth template $T$ on average---achieving at least $97.9\%$ coverage on every test instance---while occupying $64.6\%$ of the image area. For comparison with a contrast-based localizer, we construct an analogous mask $R$ from $\Delta h_{\tilde\theta}$, the stronger of the two curvature contrasts of \citet{kimlee2026} on the development split; this mask covers $97.6\%$ of the template on average, but falls below $99\%$ coverage on $9.4\%$ of the test instances.

\paragraph{Mitigation methods.} We evaluate three prompt mitigations $\tilde y$ using the hyperparameter settings of their original papers or official repositories. First, RTA~\cite{somepalli2023understanding} inserts $4$ random tokens into the prompt at the default setting and $8$ at the strong setting. Second, for \citet{wen2024detecting}, we optimize the prompt embedding using Adam with learning rate $0.05$ for up to $20$ steps, stopping once the first-step text-conditional noise norm falls to $30\%$ of its initial value ($15\%$ at the strong setting). Third, for Bright Ending~\cite{chen2025bright}, we follow the authors' official implementation: we optimize the prompt embedding (keeping the start token fixed) using AdamW with learning rate $0.05$ for up to $10$ steps, minimizing the $\ell_2$ norm of the masked text-conditional noise difference divided by the spatial mean of the mask until this normalized objective drops below a fixed threshold of $1$. Because the objective is normalized by the mean of the mask, different masks reach the fixed stopping threshold of $1$ at different rates: within $10$ steps at the original initial noise $\x{x}_1$, the threshold is reached on $97.1\%$ of test instances with the Bright Ending map (computed as in App.~\ref{supp:published}), $97.8\%$ without a mask, and $86.0\%$ with the JaRS mask $R$. Because \citet{chen2025bright} calibrate this threshold of $1$ specifically for SD v1.4, we evaluate Bright Ending on SD v1.4 only.

\paragraph{Evaluation metrics.} We report five complementary metrics. `Copied (MSE)' is the fraction of mitigated images that still reproduce the ground-truth template under the benchmark's copy criterion~\cite{webster2023reproducible}, defined as a mean squared error below $2{,}500$ to the template image inside the template mask at $256 \times 256$ resolution. `Copied (SSCD)' measures the copy rate using self-supervised copy-detection descriptors~\cite{pizzi2022self}, counting an instance as copied when the cosine similarity between the mitigated and template images (with pixels outside the template mask zeroed out) exceeds $0.5$. To quantify background preservation, LPIPS$_{\bar R}$ and LPIPS$_{\bar T}$ measure the perceptual distance~\cite{zhang2018unreasonable} between the mitigated image and the original unmitigated generation, averaged over the pixels outside the mask $R$ and outside the ground-truth template $T$, respectively. Whereas LPIPS$_{\bar R}$ is evaluated on the complement of each method's mask $R$, LPIPS$_{\bar T}$ is evaluated over the exact same background pixels for every method. Note that under our procedure, even though the latent outside $R$ is identical to $\xo$, LPIPS$_{\bar R}$ is slightly above zero because the autoencoder's convolutional decoder mixes latent features across the boundary of $R$. Finally, CLIP measures semantic alignment between the full mitigated image and the text prompt~\cite{radford2021learning} using OpenCLIP ViT-B/32.

\begin{table}
    \centering
    \small
    \resizebox{\columnwidth}{!}{%
    \begin{tabular}{@{}llccccc@{}}
        \toprule
        $\tilde y$ & Sampling & \begin{tabular}[c]{@{}c@{}}Copied\\(MSE) $\downarrow$\end{tabular} & \begin{tabular}[c]{@{}c@{}}Copied\\(SSCD) $\downarrow$\end{tabular} & LPIPS$_{\bar R}$ $\downarrow$ & LPIPS$_{\bar T}$ $\downarrow$ & CLIP $\uparrow$ \\
        \midrule
        \multirow{4}{*}{RTA} & Whole image & 0.245 & 0.356 & 0.600 & 0.591 & 21.3 \\
        & Whole, fresh noise & 0.299 & 0.327 & 0.651 & 0.644 & 21.6 \\
        & Region $R$ (ours) & 0.284 & 0.306 & 0.121 & 0.188 & 24.8 \\
        & Region $R$, $\Delta h_{\tilde\theta}$ & 0.306 & 0.345 & 0.111 & 0.195 & 24.8 \\
        \midrule
        \multirow{4}{*}{Wen et al.} & Whole image & 0.101 & 0.112 & 0.799 & 0.777 & 18.3 \\
        & Whole, fresh noise & 0.104 & 0.140 & 0.863 & 0.837 & 16.6 \\
        & Region $R$ (ours) & 0.058 & 0.072 & 0.185 & 0.276 & 20.5 \\
        & Region $R$, $\Delta h_{\tilde\theta}$ & 0.068 & 0.115 & 0.180 & 0.300 & 20.9 \\
        \bottomrule
    \end{tabular}}
    \caption{\textbf{Mitigation at the strong setting on the $278$ test TV instances of Stable Diffusion v1.4}: $8$ random tokens for RTA~\cite{somepalli2023understanding}, and the objective of \citet{wen2024detecting} reduced to $15\%$ of its initial value. Columns and rows follow \cref{tab:mitig}.}
    \label{tab:mitig:strong}
\end{table}

\paragraph{Does restricting the mitigation to $R$ help?} We first address this question on SD v1.4 (\cref{tab:mitig}). For \citet{wen2024detecting}, restricting the mitigation to $R$ improves every metric simultaneously over whole-image mitigation: the MSE copy rate drops from $20.5\%$ to $12.2\%$ ($23.4\%$ to $14.7\%$ under SSCD), background change drops by more than three-fold from $0.708$ to $0.147$ in LPIPS$_{\bar R}$ and from $0.689$ to $0.224$ in LPIPS$_{\bar T}$, and the CLIP score rises from $20.4$ to $24.0$. For RTA, restricting sampling to $R$ maintains the same MSE copy rate ($43.2\%$) and lowers the SSCD copy rate from $59.0\%$ to $47.5\%$, while cutting LPIPS$_{\bar T}$ from $0.437$ to $0.148$ and raising CLIP from $25.0$ to $26.8$. These gains persist at the strong mitigation setting (\cref{tab:mitig:strong}): region-restricted \citet{wen2024detecting} lowers the copy rate from $10.1\%$ to $5.8\%$ while reducing LPIPS$_{\bar T}$ from $0.777$ to $0.276$, and region-restricted RTA reduces LPIPS$_{\bar T}$ from $0.591$ to $0.188$ with comparable copy rates ($28.4\%$ vs.\ $24.5\%$ MSE, and $30.6\%$ vs.\ $35.6\%$ SSCD). 

Furthermore, if the prompt embedding of \citet{wen2024detecting} is optimized at the original memorization-triggering initial noise $\x{x}_1$ rather than at $\x{x}_1^R$ (final row of the \citeauthor{wen2024detecting} block in \cref{tab:mitig}), the stronger initial guidance spike at $\x{x}_1$ drives a larger update to the prompt embedding, lowering the copy rate even further to $9.0\%$ with an LPIPS$_{\bar R}$ of $0.158$ (compared to $20.5\%$ and $0.708$ for whole-image mitigation), at the cost of a slightly lower CLIP score ($22.2$ vs.\ $24.0$). Throughout the remaining rows, we optimize at $\x{x}_1^R$ so that every method calibrates its embedding at the latent from which it samples.

Two ablation controls in \cref{tab:mitig} confirm that both prompt mitigation and fresh noise inside $R$ are essential. First, redrawing fresh noise inside $R$ while keeping the original unmitigated prompt still leaves $69.4\%$ of images copied, whereas replacing the prompt with the null prompt inside $R$ eliminates copying completely but collapses prompt alignment to a CLIP score of $13.6$; prompt mitigation is therefore required inside $R$. Second, running whole-image mitigation from the exact same fresh noise realization $\x{\xi}$ leaves $24.8\%$ of images copied under \citet{wen2024detecting} and $47.1\%$ under RTA, while altering the entire image ($0.761$ and $0.529$ LPIPS$_{\bar T}$). Conversely, omitting fresh noise inside $R$ during region-restricted sampling (`\emph{Region $R$, no fresh noise}') raises the copy rate of \citet{wen2024detecting} from $12.2\%$ to $34.5\%$, showing that redrawing the noise inside $R$ is necessary to break the seed-specific bias toward the memorized template.

\begin{table}
    \centering
    \small
    \resizebox{\columnwidth}{!}{%
    \begin{tabular}{@{}llccccc@{}}
        \toprule
        $\tilde y$ & Sampling & \begin{tabular}[c]{@{}c@{}}Copied\\(MSE) $\downarrow$\end{tabular} & \begin{tabular}[c]{@{}c@{}}Copied\\(SSCD) $\downarrow$\end{tabular} & LPIPS$_{\bar R}$ $\downarrow$ & LPIPS$_{\bar T}$ $\downarrow$ & CLIP $\uparrow$ \\
        \midrule
        None & Original image & 0.998 & 1.000 & 0.000 & 0.000 & 30.5 \\
        \midrule
        \multirow{3}{*}{RTA} & Whole image & 0.832 & 0.868 & 0.170 & 0.168 & 30.2 \\
        & Whole, fresh noise & 0.091 & 0.100 & 0.451 & 0.482 & 31.1 \\
        & Region $R$ (ours) & 0.091 & 0.106 & 0.122 & 0.200 & 31.1 \\
        \midrule
        \multirow{3}{*}{Wen et al.} & Whole image & 0.265 & 0.478 & 0.382 & 0.395 & 27.5 \\
        & Whole, fresh noise & 0.015 & 0.068 & 0.546 & 0.566 & 29.7 \\
        & Region $R$ (ours) & 0.025 & 0.096 & 0.135 & 0.214 & 28.7 \\
        \bottomrule
    \end{tabular}}
    \caption{\textbf{Mitigation on the $471$ test TV instances of Stable Diffusion v2.1} at the default setting. Columns and rows follow \cref{tab:mitig}. The copy metrics compare against each instance's assigned template; on this benchmark, mitigated generations frequently switch to a different memorized template from the same product family (see text).}
    \label{tab:mitig:v2}
\end{table}

\paragraph{Where should the mask act: in the loss or in the sampler?} Comparing against Bright Ending~\cite{chen2025bright} isolates the effect of applying a spatial mask in the optimization loss versus in the sampler (\cref{tab:mitig}). Under whole-image sampling, adding a spatial mask to the loss reduces the copy rate from $13.7\%$ (unmasked) to $10.1\%$ (JaRS mask $R$) and $10.8\%$ (Bright Ending map), but fails to preserve the unmemorized background: LPIPS$_{\bar T}$ remains between $0.680$ and $0.694$ across all three rows. A mask in the loss therefore cannot prevent background alteration when the entire image is regenerated. Conversely, applying our region-restricted sampler reduces LPIPS$_{\bar T}$ to $0.188$--$0.204$, but pairing it with the Bright Ending optimization objective yields higher copy rates of $19.1\%$--$21.6\%$. This mismatch stems from Bright Ending's fixed absolute stopping threshold of $1$: because replacing the memorization-triggering noise inside $R$ to form $\x{x}_1^R$ lowers the initial objective from a median of $4.5$ (at $\x{x}_1$) to $2.8$ (at $\x{x}_1^R$), the optimization reaches the fixed threshold of $1$ in fewer steps ($4.1$ vs.\ $5.6$ on average) and under-optimizes the embedding. In contrast, the relative stopping criterion of \citet{wen2024detecting} (reducing the objective to $30\%$ of its initial value at $\x{x}_1^R$) adapts automatically to the starting latent, achieving a $12.2\%$ copy rate ($9.0\%$ when optimized at $\x{x}_1$) while reducing background change outside the template by over three-fold compared to published Bright Ending ($0.224$ vs.\ $0.680$ LPIPS$_{\bar T}$) and improving CLIP alignment ($24.0$ vs.\ $22.8$).

\paragraph{Which localization mask works better?} Replacing the JaRS mask $R$ with a mask constructed identically from the checkpoint contrast $\Delta h_{\tilde\theta}$ yields comparable background preservation (last block of \cref{tab:mitig} and \cref{tab:mitig:strong}), but JaRS achieves consistently lower copy rates without requiring a reference checkpoint. Specifically, when paired with RTA, the JaRS mask lowers the MSE copy rate from $47.1\%$ to $43.2\%$ at the default setting and from $30.6\%$ to $28.4\%$ at the strong setting, consistent with its higher template coverage ($99.9\%$ vs.\ $97.6\%$). When paired with \citet{wen2024detecting}, the two masks achieve similar MSE copy rates ($12.2\%$ at default; $5.8\%$ vs.\ $6.8\%$ at strong), while the JaRS mask reduces the SSCD copy rate from $17.3\%$ to $14.7\%$ at the default setting and from $11.5\%$ to $7.2\%$ at the strong setting.

\paragraph{Generalization to Stable Diffusion v2.1.} On SD v2.1 (\cref{tab:mitig:v2}), template copying is heavily governed by the initial noise realization: whole-image RTA starting from the original noise $\x{x}_1$ leaves $83.2\%$ of images copied, whereas starting from fresh noise reduces the copy rate to $9.1\%$. Our region-restricted procedure matches the low copy rates of fresh-noise whole-image mitigation ($9.1\%$ for RTA, and $2.5\%$ vs.\ $1.5\%$ for \citet{wen2024detecting}) while reducing perceptual change outside the template by more than half ($0.200$ vs.\ $0.482$ LPIPS$_{\bar T}$ for RTA, and $0.214$ vs.\ $0.566$ for \citet{wen2024detecting}). Thus, on SD v2.1, the primary advantage of region-restricted mitigation is preserving the unmemorized background while retaining the copy reduction of fresh-noise sampling. We also note an important caveat regarding the copy metrics on SD v2.1: many memorized prompts in this benchmark are product captions of Shaw Floors carpets~\cite{kimlee2026}, and mitigating a prompt frequently causes the model to switch to a different memorized carpet template from the same family. When we evaluate each mitigated image against all $57$ templates in the benchmark rather than only its assigned instance template, the MSE copy rate rises from $2.5\%$ to $15.5\%$ for region-restricted \citet{wen2024detecting} (and from $1.5\%$ to $38.9\%$ for whole-image fresh-noise \citet{wen2024detecting}), and from $9.1\%$ to $73.7\%$ for region-restricted RTA ($83.4\%$ for whole-image fresh-noise RTA). Standard per-instance copy rates therefore understate residual template memorization on SD v2.1 across all methods, and even under all-template evaluation our region-restricted procedure copies less often than whole-image mitigation.

\section{Relation to TACO} \label{supp:taco}

TACO~\cite{gu2026tacotrajectoryawarecurvatureobservation} is a contrast-based localization method that builds on the null-prompt contrast of \citet{kimlee2026}. It estimates the diagonal curvature of the normalized difference between the conditional and unconditional noise predictions, $(\x{\epsilon}_{\x{\theta}}(\xt, t, y) - \x{\epsilon}_{\x{\theta}}(\xt, t, \varnothing)) / \sigma_t$, using $16$ structured Walsh-Hadamard probes at the last two sampling steps ($48$ and $49$ of $50$). The resulting map is then refined by sigmoid reweighting, latent-guided edge-aware smoothing, and a weighted average over the two steps. Like $\Delta h_\varnothing$, TACO therefore requires a null-prompt evaluation and applies only to text-conditional models. In contrast, JaRS requires no contrast prompt or reference checkpoint, and also applies to unconditional models such as our DiT. TACO is not evaluated on prompt-level detection.

TACO is evaluated on the benchmark of \citet{kimlee2026}, but on all prompts rather than a held-out test split, and its mixed setting adds non-memorized images without the class balancing we follow. Because the evaluation protocols differ, the numbers reported for TACO are not comparable to those in \cref{tab:main}.

TACO also addresses mitigation, but in a different way from App.~\ref{supp:mitig}. It uses its map as a soft spatial weight on the updates of existing whole-image mitigation methods, namely initial-noise adjustment and early-stage perturbations of classifier-free guidance, and evaluates the trade-off between SSCD similarity and CLIP score or ImageReward on the prompts of \citet{webster2023reproducible}. In contrast, we restrict prompt-based mitigations, such as the prompt-embedding optimization of \citet{wen2024detecting}, to the region localized by JaRS, keep the latent outside this region fixed during sampling, and measure the copy rate and the perceptual change outside the region. Because the base mitigation methods and the evaluation metrics differ, the two results are not directly comparable.

\end{document}